\documentclass[runningheads]{llncs}

\usepackage[mobile]{eccv}

\usepackage{eccvabbrv}

\usepackage{mathtools}
\usepackage{graphicx}
\usepackage{booktabs}
\usepackage{enumitem}
\usepackage[font=small]{caption}
\usepackage[accsupp]{axessibility}  %

\usepackage{hyperref}

\usepackage{orcidlink}

\usepackage{multirow}
\newcommand{\mypar}[1]{\vspace{2mm}\noindent\textbf{#1}}

\newcommand{\x}[0]{{\mathbf x}}

\begin{document}

\title{Factorized Diffusion: Perceptual\\ Illusions by Noise Decomposition}

\author{Daniel Geng* \and
Inbum Park* \and
Andrew Owens}

\authorrunning{D. Geng et al.}

\institute{University of Michigan\\ 
\email{dgeng@umich.edu} \\
\vspace{1em}\url{https://dangeng.github.io/factorized_diffusion/}
}

\maketitle

\begin{abstract}

Given a factorization of an image into a sum of linear components, we present a zero-shot method to control each individual component through diffusion model sampling. For example, we can decompose an image into low and high spatial frequencies and condition these components on different text prompts. This produces hybrid images, which change appearance depending on viewing distance. By decomposing an image into three frequency subbands, we can generate hybrid images with three prompts. We also use a decomposition into grayscale and color components to produce images whose appearance changes when they are viewed in grayscale, a phenomena that naturally occurs under dim lighting. And we explore a decomposition by a motion blur kernel, which produces images that change appearance under motion blurring. Our method works by denoising with a composite noise estimate, built from the components of noise estimates conditioned on different prompts. We also show that for certain decompositions, our method recovers prior approaches to compositional generation and spatial control. Finally, we show that we can extend our approach to generate hybrid images from real images. We do this by holding one component fixed and generating the remaining components, effectively solving an inverse problem.

\end{abstract}

\section{Introduction}
\label{sec:intro}
\begin{figure}[!t]
    \centering
    \includegraphics[width=\linewidth]{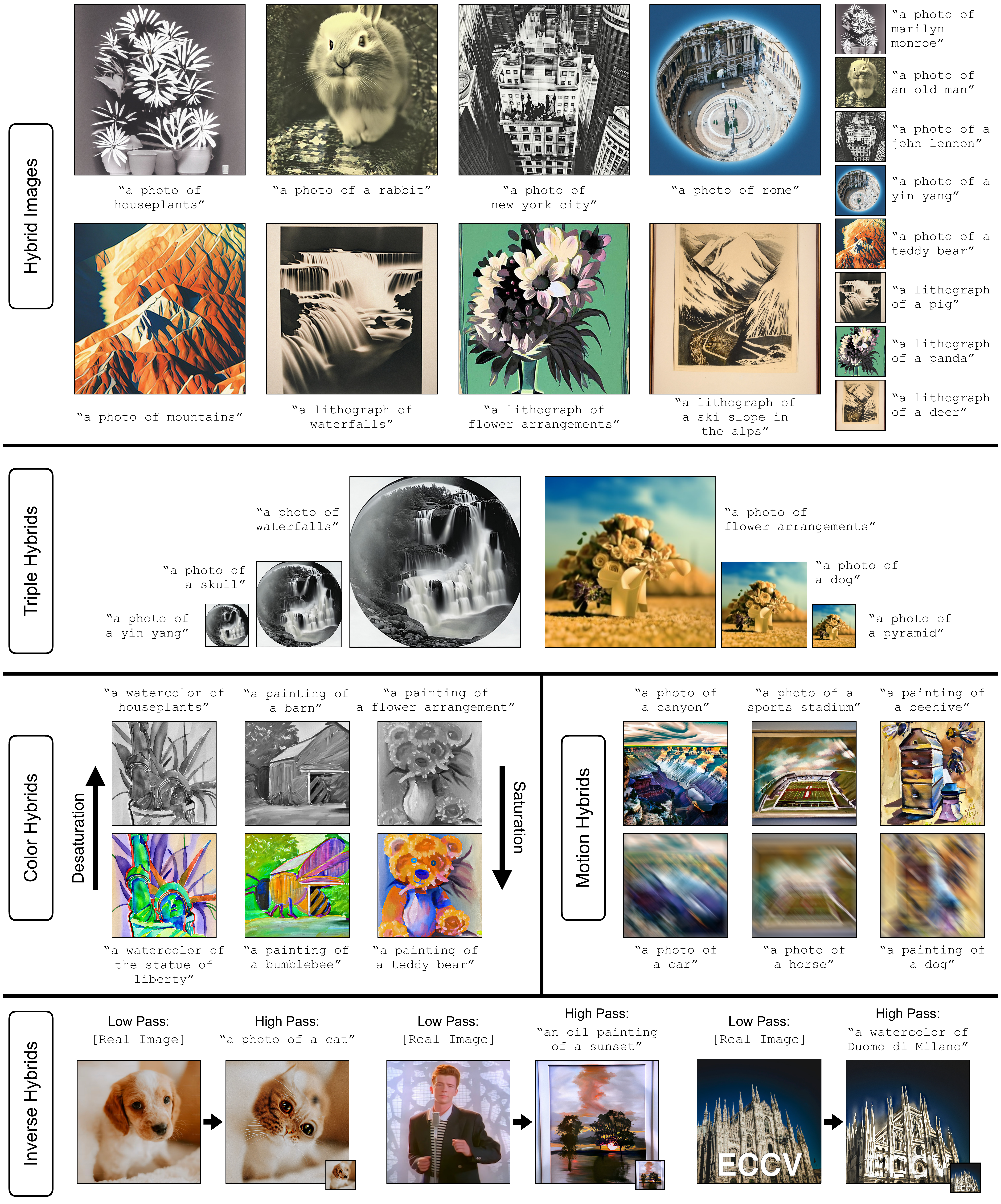} %
    \caption{{\bf Illusions by Factorized Diffusion.} By conditioning the components of a generated image with different prompts, we can use off-the-shelf text-conditioned image diffusion models to synthesize hybrid images~\cite{oliva2006hybrid}, hybrid images containing three objects, and new perceptual illusions which we refer to as {\it color hybrids} and {\it motion hybrids}, which change appearance when color is added or motion blur is induced. In addition, we can extract a component from an existing image and generate the missing components, allowing us to produce hybrid images from real images, which we term {\it inverse hybrids}. Examples shown are hand-picked. For random samples please see \cref{fig:random} and \cref{fig:sup_random}. For the hybrid images, we include insets to aid in visualization. However, perception of this effect depends on the resolution of the images, so \textbf{\textit{we highly encourage the reader to zoom so that an image fills the screen completely, or visit our \href{https://dangeng.github.io/factorized_diffusion/}{webpage} for easier viewing.}}}
    \vspace{-15mm}
\label{fig:teaser}
\end{figure}

The visual world is full of phenomena that can be understood through image decompositions. For instance, objects look blurry when seen from a distance, while up close their details are highly salient---two distinct perspectives that can be captured by a decomposition in frequency space~\cite{schyns1994blobs,oliva2006hybrid}. During the day, we see in full color, while in dim light we perceive only luminance---an effect that can be appreciated with a color space decomposition.

We present a simple method for controlling the factors of such decompositions, allowing a user to generate images that are perceived differently under different viewing conditions, yet still globally coherent. We apply this approach to generating a variety of perceptual illusions (\cref{fig:teaser}). (i) Inspired by the classic work of Oliva~\etal~\cite{oliva2006hybrid} we generate {\em hybrid images}, whose interpretation changes with viewing distance, which we achieve by controlling the generated image's low and high frequency components. A decomposition into three subbands allows us to produce hybrid images with three different prompts, which we refer to as {\em triple hybrids}. (ii) We generate color images whose appearance changes when they are viewed in grayscale, a phenomena that naturally occurs under dim lighting. We call these {\em color hybrids}, and make these by controlling image luminance separately from color. (iii) Finally, we produce images that change appearance under motion blur, by using a blur kernel to decompose an image. We refer to these as {\em motion hybrids}.

Our approach consists of a simple change to the sampling procedure of an off-the-shelf diffusion model. %
Given an image decomposition and a text prompt to control each component,
in each step of the reverse diffusion process we estimate the noise multiple times: once for each component, conditioned on its corresponding text prompt.
We then assemble a composite noise estimate by combining components from each individual noise estimate, obtained by applying the decomposition directly to the noise estimates (\cref{fig:method}).
Notably, our approach does not require finetuning~\cite{zhang2023adding,ruiz2022dreambooth} or access to auxiliary networks, as in guidance based methods~\cite{nichol2021glide,bansal2023ugd,lee2023syncdiffusion, lee2023syncdiffusion,gu2023filtered,geng2024motion}.

We also show that we can take components from existing images, and generate the remaining components conditioned on text. This recovers a simple method to solve inverse problems, and is highly related to prior work on using diffusion models for solving inverse problems~\cite{song2021scorebased, kawar2022denoising, chung2022diffusion, 
chung2022come, chung2022improving, wang2022zero, lugmayr2022repaint, song2021solving, choi2021ilvr, avrahami2022blended}. We apply this technique to producing hybrid images from real images. Finally, we show that using certain decompositions with our method recovers prior techniques for spatial~\cite{bar2023multidiffusion} and compositional~\cite{liu2021learning} control over text prompts.

\vspace{2mm}
\noindent In summary, our contributions are as follows:
\begin{itemize}[label=$\bullet$,itemsep=0pt,topsep=3pt,leftmargin=4mm]
    \item Given a decomposition of an image into a sum of components, we propose a zero-shot adaptation of diffusion models to control these components during image generation.
    \item Using our method, we produce a variety of perceptual illusions, such as images that change appearance under different viewing distances ({\it hybrid images}), illumination conditions ({\it color hybrids}), and motion blurring ({\it motion hybrids}). Each of these illusions corresponds to a different image decomposition.
    \item We provide quantitative evaluations comparing our hybrid images to those produced by traditional methods, and show that our results are better.
    \item We give an analysis and intuition for how and why our method works.
    \item We show a simple extension of our method allows us to solve inverse problems, and we apply this approach to synthesizing hybrid images from real images.
\end{itemize}

\section{Related Work}
\label{sec:relatedwork}

\mypar{Diffusion models.}
Diffusion models~\cite{sohldickstein2015diffusion,ho2020denoising,song2021scorebased,dhariwal2021diffusion,song2020denoising} are trained to denoise data corrupted by added Gaussian noise. This is achieved by estimating the noise in noisy data, potentially with some additional conditioning, such as with text embeddings. To sample data from a diffusion model, pure Gaussian noise is iteratively denoised until a clean image remains. Each denoising step consists of an update that removes a portion of the predicted noise from the noisy image, such as DDPM~\cite{ho2020denoising} or DDIM~\cite{song2020denoising}. One noteworthy application of diffusion models is for text-conditional image generation~\cite{nichol2021glide,rombach2022ldm,deepfloyd2023,saharia2022imagen}, which we build our method on top of.

\mypar{Diffusion model control.} Diffusion models are capable of both generating and editing images conditioned on text prompts. By modifying the reverse process~\cite{meng2022sdedit,geng2024visual,bar2023multidiffusion,zhang2023diffcollage,wang2023generativepowers}, finetuning~\cite{zhang2023adding,ruiz2022dreambooth}, performing text inversion~\cite{song2020denoising, mokady2022nulltext, huberman2023edit, wallace2023edict, wu2022unifying}, swapping attention maps~\cite{hertz2022prompt,epstein2023selfguidance,Tumanyan_2023_CVPR}, supplying instructions~\cite{brooks2022instructpix2pix}, or using guidance~\cite{epstein2023selfguidance, parmar2023zeroshot,lee2023syncdiffusion,bansal2023ugd,nichol2021glide,gu2023filtered}, modifying the style, location, and appearance of content in an image has become a relatively accessible task. Another line of work on compositional generation~\cite{du2020compositional, liu2022compositional, bar2023multidiffusion,wang2023generativepowers} shows that diffusion models can generate images that conform to compositions of text prompts. Our work builds upon this, and shows that similar techniques can be applied to prompting individual components of an image to produce perceptual illusions. Our work is also similar to Wang~\etal~\cite{wang2023generativepowers}, in which a diffusion model is used to generate multiple images that blend seamlessly into a zooming video. However, we focus on generating only single images that can be understood at multiple resolutions. %

Another line of work uses pretrained diffusion models to solve inverse problems, such as colorization or inpainting, in a zero-shot setting~\cite{song2021scorebased, kawar2022denoising, chung2022diffusion, 
chung2022come, chung2022improving, wang2022zero, lugmayr2022repaint, song2021solving, choi2021ilvr, avrahami2022blended}. We show that an extension of our method recovers these techniques, and we use it to generate hybrid images, which has not been considered before.

\mypar{Computational optical illusions.}\label{sec:related_work_computational_illusions}
Optical illusions are entertaining, but can also serve as windows into human and machine perception~\cite{hertzmann2020visual,wang2020toward,gomez2019convolutional,ngo2023clip,jaini2023intriguing,szegedy2013intriguing,goodfellow2014explaining,elsayed2018adversarial}. As such, much work has gone into developing computational methods for generating optical illusions~\cite{freeman1991motion,owens2014camouflaging,guo2023ganmouflage,chandra2022designing,oliva2006hybrid,geng2024visual,tancik2023illusions,ugleh2023spiral,burgert2023illusions,chu2010camouflage}. In classic work, Oliva \etal introduced hybrid images~\cite{oliva2006hybrid}, which are images that change their appearance depending on viewing distance or duration~\cite{schyns1994blobs}. These images work by combining low frequencies from one image with high frequencies of another, exploiting the multiscale processing of human perception~\cite{oliva1997stimuli, schyns1999faces}. By contrast, our approach generates hybrid images from scratch with a diffusion model. This avoids manual alignment steps and leads to higher quality illusions.

Artists and researchers have recently used text-conditioned image diffusion models to generate optical illusions. For example, a pseudonymous artist~\cite{ugleh2023spiral} adapted a QR code generation model~\cite{qrmonster2023, zhang2023adding} to create images that subtly match a target template. While these are also images with multiple interpretations, they are restricted to binary mask templates and require a specialized finetuned model. Burgert~\etal~\cite{burgert2023illusions} use score distillation sampling~\cite{poole2022dreamfusion} to generate images that match other prompts when viewed from different orientations or overlaid on top of each other. Other methods such as Tancik~\cite{tancik2023illusions} and Geng \etal~\cite{geng2024visual} use off-the-shelf diffusion models~\cite{rombach2022ldm, deepfloyd2023} to generate multi-view optical illusions that change appearance upon transformations such as rotations, flips, permutations, skews, and color inversions. These methods work by transforming the noisy image multiple ways during the reverse diffusion process, denoising each transformed version, then averaging the noise estimates together. However, many types of transformations, like the multiscale processing considered in hybrid images, fail because they perturb the noise distribution~\cite{geng2024visual}. Like these approaches, our work also changes the reverse diffusion process to produce images that have multiple interpretations. However, our approach manipulates the {\em noise estimate} rather than the noisy image, enabling us to handle illusions that prior work cannot. Please see \cref{sec:illusion_baselines} for additional discussion and results.

\section{Method}
\label{sec:method}

For a given decomposition of an image into components, our method allows for control of each of these components through text conditioning. We achieve this by modifying the sampling procedure of a text-to-image diffusion model.

\subsection{Preliminaries: Diffusion Models}

Diffusion models sample from a distribution by iteratively denoising noisy data.  Over $T$ timesteps, they denoise pure random Gaussian noise, $\x_T$, until a clean image, $\x_0$, is produced at the final step. At intermediate timesteps, a variance schedule is followed such that the noisy image at timestep $t$ is of the form
\begin{equation}
\x_t = \sqrt{\alpha_t} \x_0 + \sqrt{1 - \alpha_t} \epsilon,
\end{equation}
where $\epsilon \sim \mathcal{N}(0, \mathbf{I})$ is a sample from a standard Gaussian distribution, and $\alpha_t$ is a predetermined variance schedule. To sample $\x_{t-1}$ from $\x_t$ the diffusion model, $\epsilon_\theta(\cdot,\cdot, \cdot)$, predicts the noise in $\x_t$, conditioned on the timestep $t$ and optionally on context $y$, such as a text prompt embedding. Afterwards, an update step, $\texttt{update}(\cdot, \cdot)$, is applied which removes a portion of the estimated noise, $\epsilon_\theta \coloneq \epsilon_\theta(\x_t, y, t)$, from the noisy image $\x_t$. The exact implementation of this step depends on the specifics of the method used, but it is---critically for our method---often a linear combination of $\x_t$ and $\epsilon_\theta$ (and possibly noise, $\mathbf{z} \sim \mathcal{N}(0, \mathbf{I}))$. For example, DDIM~\cite{song2020denoising} (with $\sigma_t=0$) performs the update as:
\begin{equation}
\x_{t-1} = \texttt{update}(\x_t, \epsilon_\theta) = \sqrt{\alpha_{t-1}}\left( \frac{\x_t - \sqrt{1-\alpha_t}\epsilon_\theta}{\sqrt{\alpha_t}} \right) + \sqrt{1 - \alpha_{t-1}}\epsilon_\theta.
\end{equation}
\subsection{Factorized Diffusion}
\definecolor{mypink}{RGB}{255, 123, 115}
\definecolor{myblue}{RGB}{41, 177, 255}
\definecolor{myorange}{RGB}{255, 147, 0}
\begin{figure}[t]
    \centering
    \includegraphics[width=0.7\linewidth]{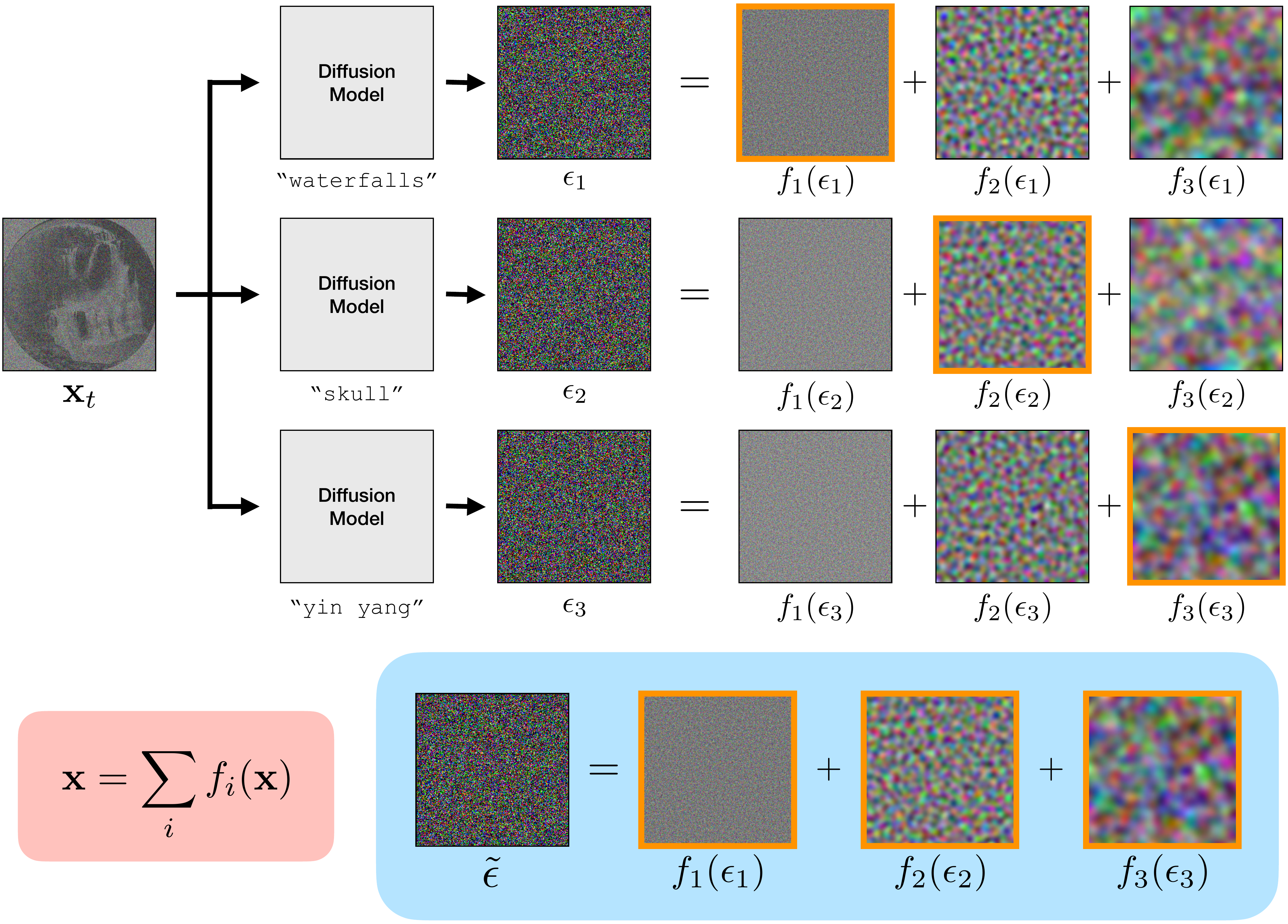}
    \caption{{\bf Factorized Diffusion.} Given an \textbf{\textcolor{mypink}{image decomposition}}, we control components of the decomposition through text conditioning during image generation. To do this, we modify the sampling procedure of a pretrained diffusion model. Specifically, at each denoising step, $t$, we construct a \textbf{\textcolor{myblue}{new noise estimate}}, $\tilde\epsilon$, to use for denoising, whose components \textbf{\textcolor{myorange}{come from components}} of $\epsilon_i$, which are noise estimates conditioned on different prompts. Here, we show a decomposition into three frequency subbands, used for creating triple hybrid images, but we consider a number of other decompositions.} \vspace{-5mm}
\label{fig:method}
\end{figure}
An overview of our method can be found in \cref{fig:method}. Our method works by manipulating the noise estimate during the reverse diffusion process such that different components of the estimate are conditioned on different prompts. Given a decomposition of an image, $\x \in \mathbb{R}^{3\times H\times W}$, into the sum of $N$ components,
\begin{equation}
\x = \sum_i^N f_i(\x),
\end{equation}
where each $f_i(\x)$ is a component, we can correspond to each component a different text prompt $y_i$. At each step of the reverse diffusion process, instead of computing a single noise estimate we compute $N$---one conditioned on each $y_i$---which we denote by $\epsilon_i = \epsilon_\theta(\x_t, y_i, t)$.
We then construct a composite noise estimate $\tilde\epsilon$ made up of components from each $\epsilon_i$:
\begin{equation}
\tilde\epsilon = \sum f_i(\epsilon_i).
\end{equation}
This new noise estimate, $\tilde\epsilon$, is used to perform the diffusion update step. In effect, each component of the image is denoised while being conditioned by a different text prompt, resulting in a clean image whose components are conditioned on the different prompts. We refer to this technique as {\em factorized diffusion}.

As noted in \cref{sec:related_work_computational_illusions}, our method is similar to recent work by Tancik~\cite{tancik2023illusions} and Geng~\etal~\cite{geng2024visual} in that we modify noise estimates with the aim of generating visual illusions. However, our method differs in that we modify only the {\em noise estimate}, and not the input to the diffusion model, $\x_t$. As a result, our method produces a different class of perceptual illusions from prior work. Please see \cref{sec:illusion_baselines} for additional discussion and results.

\subsection{Analysis of Factorized Diffusion}\label{sec:analysis}

To give intuition for why our method works, we suppose that our update function $\texttt{update}(\cdot, \cdot)$ is a linear combination of the noisy image $\x_t$ and the noise estimate $\epsilon_\theta$, as is commonly the case~\footnote{The update may also include adding random noise, $\mathbf{z}\sim\mathcal{N}(0,\mathbf{I})$, in which case our analysis still holds with a modification to the argument, discussed in \cref{sec:apdx_analysis}.}~\cite{song2020denoising,ho2020denoising}. The update function also depends on $t$, which we omit for brevity. We may then decompose the update step as
\begin{align}
    \mathbf{x}_{t-1} &= \texttt{update}(\mathbf{x}_t, \epsilon_\theta) \\ 
    &= \texttt{update}\left( \sum f_i(\mathbf{x}_t), \sum f_i(\epsilon_\theta) \right) \\ 
    &= \sum_i \texttt{update}(f_i(\mathbf{x}_t), f_i(\epsilon_\theta))\label{eq:decomp_update}
\end{align}
where the first equality is by definition of the update step, the second equality is by applying the image decomposition, and the third equality is by linearity of the update function. \cref{eq:decomp_update} tells us that an update step on $\x_t$, with $\epsilon_\theta$, can be interpreted as the {\em sum} of updates on {\em the components} of $\x_t$ and of $\epsilon_\theta$. Our method can be understood as using different conditioning on each of these components. Written explicitly, the update our method uses is
\begin{equation}
    \x_{t-1} = \sum_i^N \texttt{update}(f_i(\x_t), f_i(\epsilon_\theta(\x_t, y_i, t)).
\end{equation}
Moreover, let us write out the update step explicitly as 
\begin{equation}
    \x_{t-1} = \texttt{update}(\x_t, \epsilon_\theta) = \omega_t\x_t + \gamma_t\epsilon_\theta,
\end{equation}
for $\omega_t$ and $\gamma_t$ determined by the variance schedule and scheduler. Then if the $f_i$'s are linear, we have 
\begin{align}
    f_i(\x_{t-1}) &= f_i(\texttt{update}(\x_t, \epsilon_\theta)) \\
    &= f_i(\omega_t\x_t + \gamma_t\epsilon_\theta) \\
    &= \omega_t f_i(\x_t) + \gamma_t f_i(\epsilon_\theta) \\
    &= \texttt{update}(f_i(\x_t), f_i(\epsilon_\theta))
\end{align}
meaning that updating the $i$th component of $\x_t$ with the $i$th component of $\epsilon_\theta$ will only affect the $i$th component of $\x_{t-1}$.

\subsection{Decompositions Considered}\label{sec:decompositions_considered}

We present details for the decompositions that we consider in this paper. Results for all decompositions are presented and discussed in \cref{sec:results}.

\subsubsection{Spatial frequencies.} We consider factorizing an image into frequency subbands, and conditioning the subbands on different prompts, with the goal of producing hybrid images~\cite{oliva2006hybrid}. First, we consider a decomposition into two components:
\begin{equation}\label{eq:hybrid_decomp}
    \x = \underbrace{\x - G_\sigma(\x)}_{f_\text{high}(\x)} + \underbrace{G_\sigma(\x)}_{f_\text{low}(\x)},
\end{equation}
where $G_\sigma$ is a low pass filter implemented as a Gaussian blur with standard deviation $\sigma$, and $\x - G_\sigma(\x)$ acts as a high pass of $\x$. For a decomposition into three subbands, to make the triple hybrid images in \cref{fig:teaser}, components are levels of a Laplacian pyramid which we define as
\begin{equation}\label{eq:triple_decomp}
    \x = 
    \underbrace{\x - G_{\sigma_1}(\x)}_{f_\text{high}(\x)} +
    \underbrace{G_{\sigma_1}(\x) - G_{\sigma_2}(G_{\sigma_1}(\x))}_{f_\text{med}(\x)} +
    \underbrace{G_{\sigma_2}(G_{\sigma_1}(\x))}_{f_\text{low}(\x)}
\end{equation}
where $\sigma_1$ and $\sigma_2$ roughly define cutoffs for the low, medium, and high passes.

\subsubsection{Color spaces.}  We also consider decomposition by color space, with the goal of creating {\it color hybrids}---images with different interpretations when seen in grayscale or color. Similarly to the CIELAB color space, we decompose an image into a lightness component, {\it L}, and a chromaticity component {\em ab}. CIELAB seeks to represent colors in a perceptually uniform space, and therefore requires nonlinear transformations of RGB values. Instead, we use a simple linear decomposition. Our {\it L} component is a channel-wise average of all the pixels
\begin{equation}
f_{\text{gray}}(\x) = \frac{1}{3} \sum_{c\in \{R,G,B\}} \x_c,
\end{equation}
where $\x_c$ are the color channels of the image, $\x$, and the resultant $f_\text{gray}(\x)$ has the same shape as $\x$. We define the color component as the residual:
\begin{equation}
f_{\text{color}}(\x) = \x - f_{\text{gray}}(\x).
\end{equation}
\subsubsection{Motion blurring. } Motion blur may be modeled as a convolution with a blur kernel $\mathbf{K}$~\cite{fergus2006removing,takeda2011removing,mildenhall2018burst,brooks2019learning,yitzhaky1998direct,nayar2004motion}. To produce images that change appearance when blurred, what we call {\em motion hybrids}, we study the following decomposition:
\begin{equation}\label{eq:motion_decomp}
    \x = \underbrace{\mathbf{K}*\x}_{f_\text{motion}(\x)} + \;\; \underbrace{\x - \mathbf{K}*\x}_{f_\text{res}(\x)},
\end{equation}
where we have split an image into a motion blurred component and a residual component. We specifically study simple constant velocity motions, in which $\mathbf{K}$ may be modeled as a matrix of zeros with a line of non-zero values. This may also be thought of as decomposing an image into an oriented low frequency component, and a residual component.

\subsubsection{Spatial decomposition. } While our primary focus is on perceptual illusions, we also consider spatial masking as a decomposition. Given binary spatial masks $\mathbf{m}_i$ whose disjoint union covers the entire image, we can use the decomposition
\begin{equation}
\x = \sum_i \underbrace{\mathbf{m}_i \odot \x}_{f_i(\x)},
\end{equation}
where $\odot$ denotes element-wise multiplication and each $\mathbf{m}_i \odot \x$ is a component. The effect of this decomposition is to enable control of the prompts spatially. This is a special case of MultiDiffusion~\cite{bar2023multidiffusion}.  We discuss this connection in \cref{sec:apdx_multidiffusion}.

\subsubsection{Scaling. } A final interesting decomposition is of the form $\x = \sum^N_i a_i\x$, for $\sum^N_i a_i = 1$. Taking $a_i = \frac{1}{N}$ recovers the compositional diffusion method of Liu~\etal~\cite{liu2022compositional}, in which noise estimates are averaged to sample from conjunctions of multiple prompts. Taking $a_0 = 1-\gamma$ and $a_1 = \gamma$, gives us CFG~\cite{ho2022classifierfree} and negative prompting~\cite{negative2022}.
\vspace{-2mm}

\subsection{Inverse Problems}\label{sec:method_inverse}
If we know what one of the components must be in our generated image, perhaps extracted from some reference image $\x_\text{ref}$, we can then fix this component while generating all other components with our method. This enables us to produce hybrid images from real images (see \cref{fig:teaser,fig:inverse}). Without loss of generality, suppose we want to fix the first component. To do this, we can project $\x_t$ after every reverse process step:
\begin{equation}
    \x_t \gets f_1\left(\sqrt{\alpha_t}\x_\text{ref} + \sqrt{1 - \alpha_t}\epsilon\right) + \sum_{i=2}^N f_i(\x_t)
\end{equation}
where $\epsilon\sim\mathcal{N}(0,\mathbf{I})$, and $\alpha_t$ is determined by the variance schedule. The argument of $f_1$ is a sample from the forward process, given the reference image---that is, a noisy version of $\x_\text{ref}$ with the correct amount of noise for timestep $t$. Essentially, we project $\x_t$ such that its first component matches that of $\x_\text{ref}$. This amounts to solving a (noiseless) inverse problem characterized by $\mathbf{y} = f_1(\x$). Much work has gone into developing methods to solve inverse problems using diffusion models as priors, and this extension of our method can be viewed as a simplified version of prior work~\cite{kawar2022denoising, chung2022diffusion, wang2022zero, song2021solving, lugmayr2022repaint, choi2021ilvr, avrahami2022blended}.

\section{Results}
\label{sec:results}

We provide results organized by decomposition, followed by results on inverse problems, and then random samples. Additional implementation details can be found in \cref{sec:apdx_impl_details}, and additional results can be found in \cref{sec:apdx_more_results}.

\subsection{Hybrid Images}\label{sec:hybrid}

We show qualitative results in \cref{fig:teaser}, \cref{fig:sweep}, and \cref{fig:oliva}, as well as in \cref{fig:sup_hybrid} in the appendix. As can be seen, our method produces high quality hybrid images. Interestingly, we were also able to produce hybrid images with three different prompts (\cref{fig:teaser,fig:sup_triple}) by using the Laplacian pyramid decomposition (\cref{eq:triple_decomp}). While prior work~\cite{sripian2020hybrid} has attempted to generate these {\it triple hybrids} with traditional methods, our method far exceeds their results in terms of quality and recognizability (for details, see \cref{sec:apdx_prior_triple}).

\begin{figure}[t!]
    \centering
    \includegraphics[width=\linewidth]{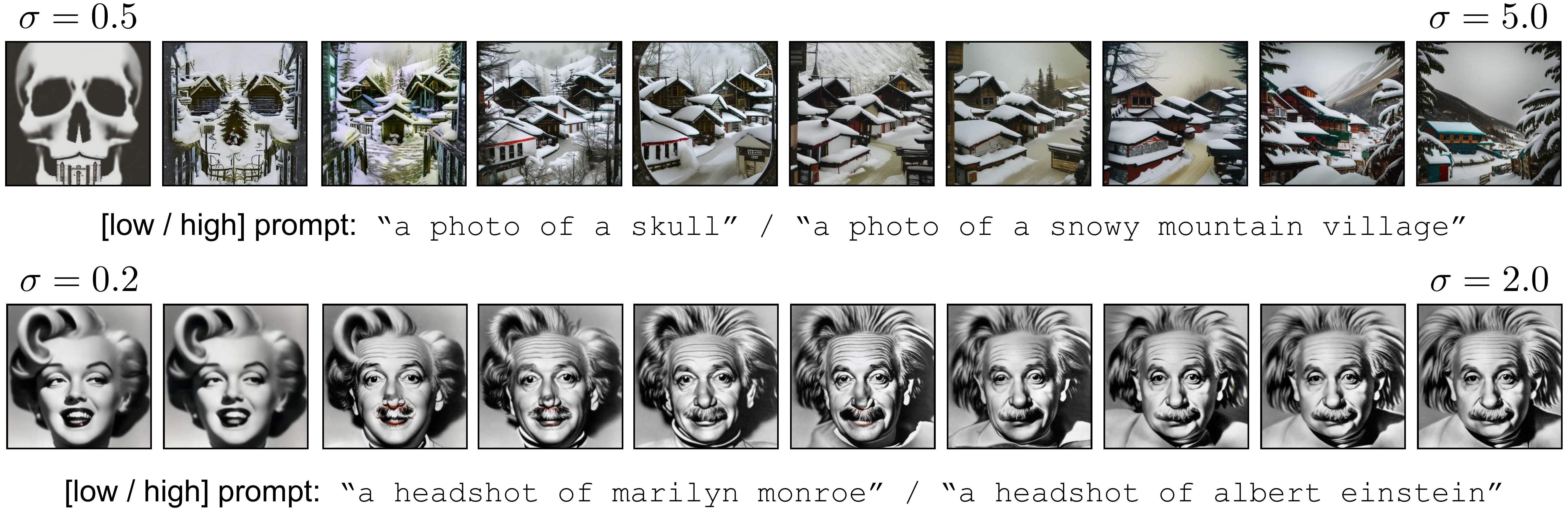}
    \caption{{\bf Effect of $\sigma$.} We show a linear sweep over the $\sigma$ value used in our hybrid decomposition. A lower $\sigma$ results in the low pass prompt being more prominent, and vice-versa. In between lies hybrid images. \textbf{\textit{Best viewed digitally, with zoom.}}}
    \vspace{-3mm}
\label{fig:sweep}
\end{figure}

\subsubsection{Effect of blur kernel. } In \cref{fig:sweep} we show how the strength of the Gaussian blur, $\sigma$, affects results. A lower $\sigma$ value corresponds to a higher cut-off frequency on the low-pass filter, and results in the low-pass prompt being more prominently featured. Interpolating between $\sigma$ values gives hybrid images.

\subsubsection{Comparisons to Oliva~\etal~\cite{oliva2006hybrid}. }

\begin{figure}[t!]
    \centering
    \includegraphics[width=\linewidth]{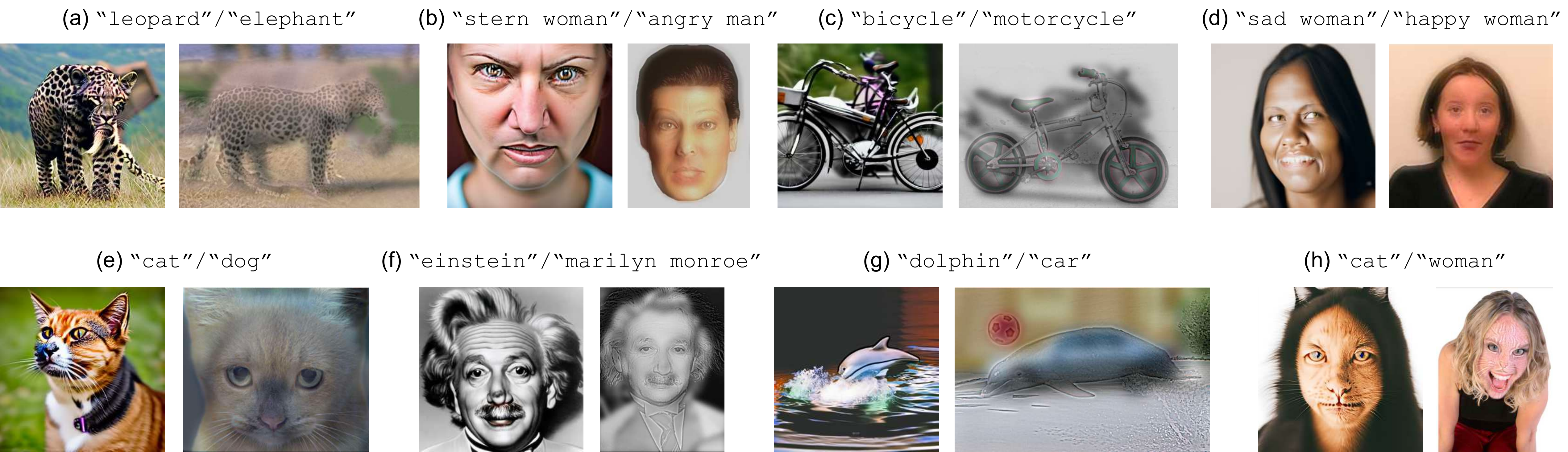}
    \caption{{\bf Comparison to Oliva~\etal~\cite{oliva2006hybrid}. } We take hybrid images from Oliva~\etal~\cite{oliva2006hybrid}, and generate our own versions. Left is from our method, and right is from Oliva~\etal's. As can be seen, our method produces much more realistic images while still containing both subjects. \textbf{\textit{Best viewed digitally, with zoom.}}}
    \vspace{-6mm}
\label{fig:oliva}
\end{figure}

In \cref{fig:oliva} we qualitatively compare our method to samples from Oliva~\etal~\cite{oliva2006hybrid}. We directly take samples from~\cite{oliva2006hybrid}, and manually create prompts to generate corresponding hybrids using our method. As can be seen, our hybrids are considerably more realistic, while containing the desired prompts at different viewing distances. One advantage our technique has is that the low frequency and high frequency components are generated with knowledge of each other, as the diffusion model is given the entire image. This is in contrast to the hybrid images of Oliva~\etal, in which frequency components are extracted from two independent images and combined. Moreover, these two images must be found and made to align manually, whereas our method simply generates low and high frequency components that align well.

\begin{table}[t!]
    \begin{center}
    \caption{{\bf Human Studies.} We compare our hybrid images and Oliva~\etal's with a two-alternative forced choice test. Participants were shown results from \cref{fig:oliva}, and were asked which images better contained the prompts, and which were of higher overall quality. Percentages denote the proportion that chose our method. Please see \cref{sec:apdx_human_impl} for additional details. We find that our method is rated as both higher in quality and better aligned with the prompts. ($N = 77$)}
    
    \label{tbl:hybrid_human}
    \setlength\tabcolsep{3pt}
    \resizebox{0.7\linewidth}{!}{
    \begin{tabular}{lccccccccc}
    \toprule
     
    & (a) & (b) & (c) & (d) & (e) & (f) & (g) & (h) & Average \\
    \midrule
    High Prompt & 70.1\% & 81.8\% & 63.6\% & 51.9\% & 61.0\% & 53.2\% & 70.1\% & 54.5\% & 63.3\% \\
    Low Prompt & 83.1\% & 84.4\% & 74.0\% & 87.0\% & 93.5\% & 87.0\% & 75.3\% & 81.8\% & 83.3\% \\
    Quality & 92.2\% & 87.0\% & 83.1\% & 77.9\% & 85.7\% & 79.2\% & 92.2\% & 33.8\% & 78.9\% \\
    \bottomrule
    \end{tabular}
   }
    \end{center}
    \vspace{-4mm}
\end{table}

\begin{table}[t!]
    \begin{center}
    \caption{{\bf Hybrid Image CLIP Evaluation.} We evaluate hybrid images by reporting the maximum clip score over different amounts of blurring. We report the max to compensate for the fact that different hybrid images may be best viewed at different resolutions. Please see \cref{fig:oliva} for the referenced hybrid images, and \cref{sec:apdx_metrics} for metric implementation details.}
    
    \label{tbl:hybrid_clip}
    \setlength\tabcolsep{3pt}
    \resizebox{0.8\linewidth}{!}{
    \begin{tabular}{llccccccccc}
    \toprule
     
    & Method & (a) & (b) & (c) & (d) & (e) & (f) & (g) & (h) & Average \\
    \midrule
    \multirow{ 2}{*}{Low Pass} & Oliva~\etal~\cite{oliva2006hybrid} & 0.268 & 0.258 & 0.316 & 0.250 & 0.237 & 0.264 & 0.257 & 0.241 & 0.261 \\
    & Ours & 0.286 & 0.252 & 0.307 & 0.273 & 0.275 & 0.260 & 0.244 & 0.269 & \textbf{0.271} \\
    \cmidrule(r){1-11}
    \multirow{ 2}{*}{High Pass} & Oliva~\etal~\cite{oliva2006hybrid} & 0.297 & 0.230 & 0.306 & 0.272 & 0.276 & 0.306 & 0.260 & 0.231 & 0.272 \\
    & Ours & 0.321 & 0.242 & 0.301 & 0.258 & 0.292 & 0.324 & 0.320 & 0.277 & \textbf{0.292} \\
    \bottomrule
    \end{tabular}
   }
    \end{center}
    \vspace{-8mm}
\end{table}

We also provide quantitative comparisons between our hybrid images and those of Oliva~\etal~\cite{oliva2006hybrid}. In \cref{tbl:hybrid_human} we show results of a two-alternative forced choice (2AFC) study, in which human participants are asked to choose between our hybrid images or Oliva~\etal's. Participants of the study were asked which image better contained the prompts, and which of the images were of higher overall quality. For details, please see \cref{sec:apdx_human_impl}. We find that participants consistently choose our images as being both higher in quality and better containing the prompts. 

Finally, we present CLIP alignment scores in \cref{tbl:hybrid_clip}. To account for the fact that the hybrid images are best viewed at many different resolutions, we report the maximum CLIP score between the prompt and the image blurred by different amounts. Please see \cref{sec:apdx_metrics} for metric implementation details. We find that our method generates hybrid images with better alignment to the prompts.

\subsection{Other Decompositions}
\subsubsection{Color hybrids.} We provide qualitative color hybrid results in \cref{fig:teaser} and \cref{fig:colorization}, as well as in \cref{fig:sup_color} in the appendix. As can be seen, the grayscale image aligns with one prompt, while the color image aligns with another. For example, in the \texttt{"rabbit"/"volcano"} image from \cref{fig:colorization} the ears of the rabbit are repurposed as plumes of lava in the grayscale image. Note that it is not sufficient to simply add arbitrary amounts of color to a grayscale image to achieve this effect, as the colors added must not change the alignment of the grayscale image with its prompt. One interesting application of this technique is to produce images that appear different under bright lighting versus dim lightning, where human vision has a much harder time discerning color.

\begin{figure}[t!]
    \centering
    \includegraphics[width=\linewidth]{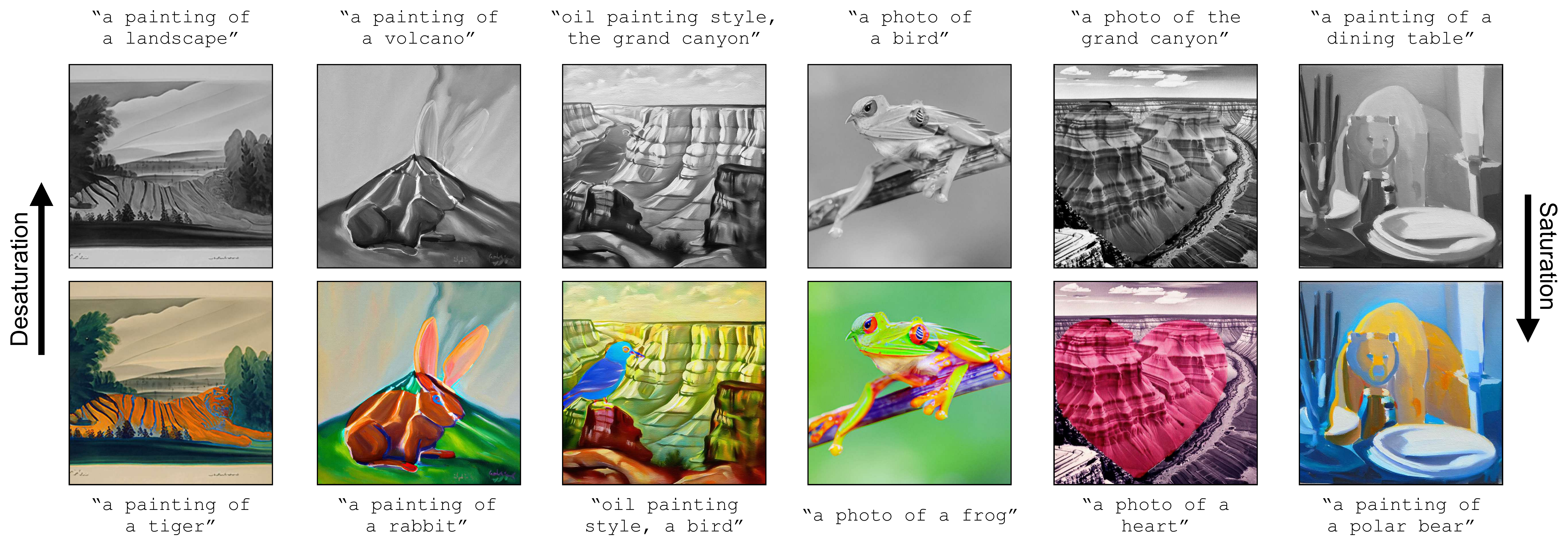}
    \caption{{\bf Color Hybrids.} We show additional {\it color hybrid} results. These are images that change appearance when color is added or subtracted away. These images change appearance when moved from bright to dim lighting, in which color is harder to see.}
    \vspace{-0.5em}
\label{fig:colorization}
\end{figure}
\begin{figure}[t!]
    \centering
    \includegraphics[width=\linewidth]{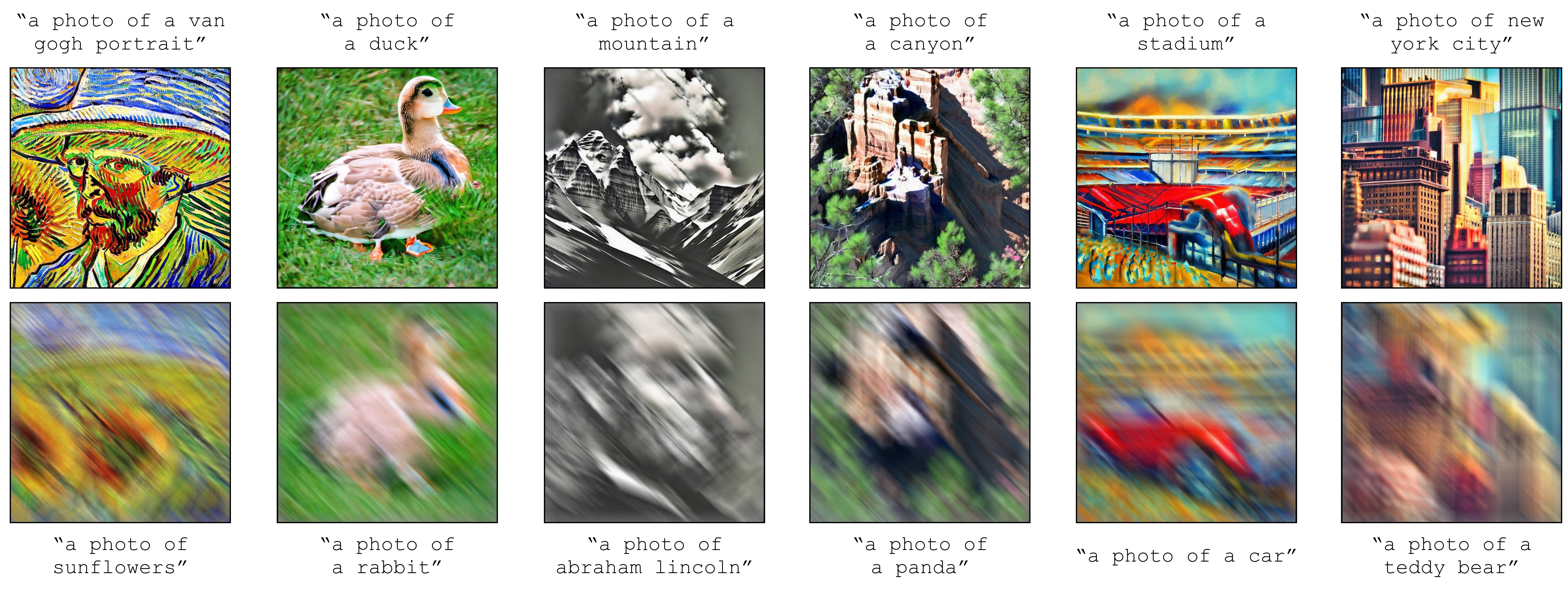}
    \caption{{\bf Motion Hybrids.} We show additional {\it motion hybrid} results. These are images that change appearance when motion blurred. Here, the motion is from upper left to bottom right.}
    \vspace{-0.5em}
\label{fig:motion_hybrids}
\end{figure}
\subsubsection{Motion hybrids}
We provide qualitative motion hybrid results in \cref{fig:teaser} and \cref{fig:motion_hybrids}, as well as in \cref{fig:sup_motion} in the appendix. These are images that change appearance when motion blurred. For all motion hybrids in the paper, we use a blur kernel of $\mathbf{K} = \frac{1}{k}\mathbf{I} \in \mathbb{R}^{k\times k}$, with $k=29$, corresponding to a diagonal motion from upper left to bottom right.

\subsubsection{Spatial decomposition}\label{sec:multidiffusion}

By decomposing an image into disjoint spatial regions and applying our method, we can recover a technique that is a special case of MultiDiffusion~\cite{bar2023multidiffusion}. Using this method, we can effect fine-grained control over where the text prompts act spatially, as shown in \cref{fig:multidiffusion}. For additional discussion, please see \cref{sec:apdx_multidiffusion}.

\subsubsection{Scaling decomposition} 
\begin{figure}[t!]
    \centering
    \includegraphics[width=\linewidth]{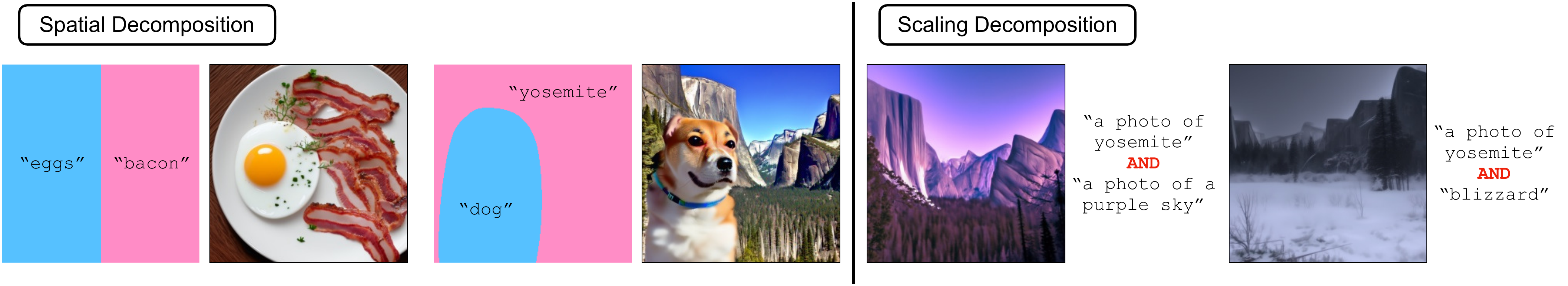}\vspace{-3mm}
    \caption{{\bf Spatial and Scaling Decompositions. } Special cases of our method reduce to prior work. {\bf (Left)} Decomposing images into spatial regions recovers a special case of MultiDiffusion~\cite{bar2023multidiffusion}, and allows us to assign prompts to spatial regions. {\bf (Right)} Decomposing an image by scaling allows us to compose concepts, and recovers the method of Liu~\etal~\cite{liu2022compositional}.}
\label{fig:multidiffusion}
\end{figure}

By using the scaling decomposition with $a_i=\frac{1}{N}$, our method reduces exactly to prior work on compositionality in diffusion models, by Liu~\etal~\cite{liu2022compositional}. Specifically, we recover the conjunction operator proposed by Liu~\etal. We demonstrate this in \cref{fig:multidiffusion}, but refer the reader to~\cite{liu2022compositional} for more examples.

\begin{figure}[t!]
    \centering
    \includegraphics[width=\linewidth]{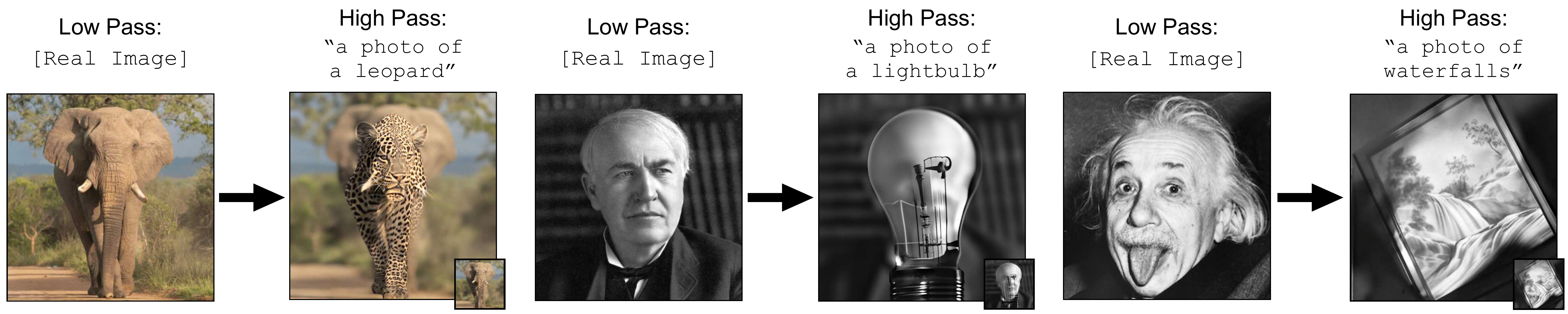}\vspace{-3mm}
    \caption{{\bf Hybrids from Real Images.} We show hybrid images generated from real images. We take low or high passes of real images, and use our method to fill in the missing component, conditioned on a prompt. \textbf{\textit{Best viewed digitally, with zoom.}}}
\label{fig:inverse}
\end{figure}

\subsection{Inverse Problems}\label{sec:inverse}
As discussed in \cref{sec:method_inverse}, we can modify our approach to solve inverse problems, resulting in a technique highly similar to prior work~\cite{wang2022zero,kawar2022denoising,chung2022come,chung2022diffusion,chung2022improving,song2021scorebased,lugmayr2022repaint, choi2021ilvr, avrahami2022blended}. While previous work investigates using diffusion priors for solving problems such as colorization, inpainting, super-resolution, or phase retrieval, we apply the idea towards generating hybrid images from real images. Specifically, we take low or high frequency components from a real image and use our method to fill in the missing components, conditioned on a prompt. Results are shown in \cref{fig:teaser} and \cref{fig:inverse}. We also provide colorization results in \cref{sec:apdx_colorization}.

\subsection{Limitations, Random Examples, and Societal Impacts}
\begin{figure}[t]
    \centering
    \includegraphics[width=\linewidth]{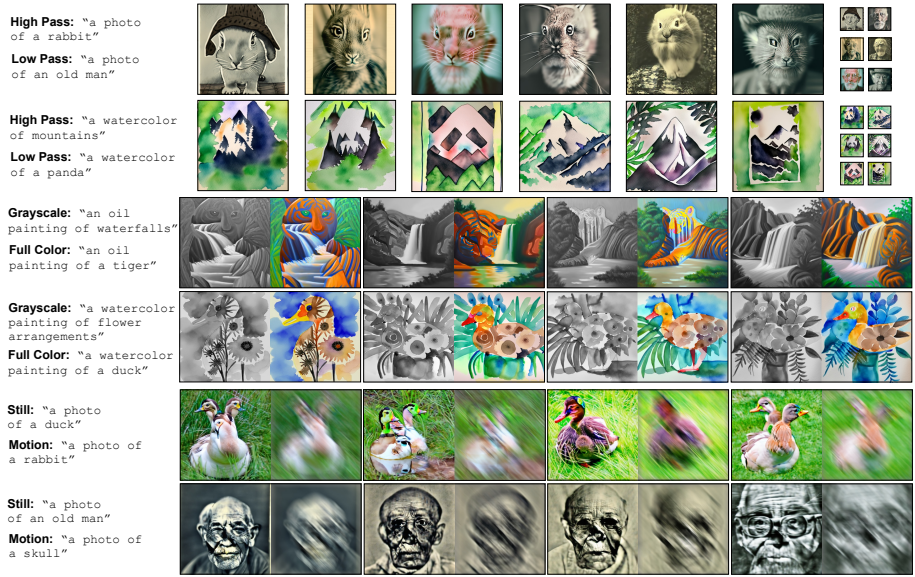}
    \caption{{\bf Random Samples.} We provide random samples for selected prompts and decompositions. As can be seen, most random results are of passable quality, with some catastrophic failures, and some very high quality illusions. More random samples are shown in \cref{fig:sup_random}.}
    \vspace{-5mm}
\label{fig:random}
\end{figure}

While our method can produce decent images fairly consistently, very high quality images are rarer. This can be seen in \cref{fig:random} and \cref{fig:sup_random}, in which we visualize random samples for hybrid images, color hybrids, and motion hybrids. We attribute this to the fact that our method produces images that are highly out-of-distribution for the diffusion model. Another failure case of our method is that the prompt for one component may dominate the generated image. Empirically, the success rate of our method can be improved by carefully choosing prompt pairs (see \cref{sec:apdx_prompts} for additional discussion), or by manually tuning decomposition parameters. Finally, our method relies of pixel diffusion models, as the image decompositions must be done in pixel space (see \cref{sec:apdx_impl_details,sec:apdx_latent} for additional discussion and results).

The ability to better control image synthesis opens numerous societal and ethical considerations. We apply our method to generating illusions, which in a sense seeks to deceive perception, possibly leading to applications in misinformation. We believe this and other concerns deserve further study.

\vspace{-2mm}

\section{Conclusion}
\label{sec:conclusion}

We present a zero-shot method that enables control over different components of an image through diffusion model sampling and apply it to the task of creating perceptual illusions. Using our method, we synthesize hybrid images, hybrid images with three prompts, and new classes of illusions such as {\it color hybrids} and {\it motion hybrids}. We give an analysis and provide intuition for why our method works. For certain image decompositions, we show that our method reduces to prior work on compositional generation and spatial control of diffusion models. Finally, we make a connection to inverse problems, and use this insight to generate hybrid images from real images.

\subsubsection{Acknowledgements.} 
We thank Patrick Chao, Aleksander Holynski, Richard Zhang, Trenton Chang, Utkarsh Singhal, Huijie Zhang, Bowen Song, Jeongsoo Park, Jeong Joon Park, Jeffrey Fessler, Liyue Shen, Qing Qu, Antonio Torralba, and Alexei Efros for helpful discussions. We also thank Walter Scheirer, Luba Elliott, and Nicole Finn for reaching out and giving us the (amazing) opportunity to create an illusion for the CVPR 2024 T-shirt as part of the AI Art Gallery (see \cref{sec:apdx_cvpr_tshirt}). Daniel is supported by the National Science Foundation Graduate Research Fellowship under Grant No. 1841052.

\bibliographystyle{splncs04}
\bibliography{main}

\clearpage
\appendix

\section{Implementation Details}\label{sec:apdx_impl_details}

\subsection{Pixel Diffusion Model}
For all experiments we use the pixel diffusion model DeepFloyd IF~\cite{deepfloyd2023}, as opposed to more common latent diffusion models. This is because the frequency subband, color space, and motion decompositions are not meaningful in latent space. For example, averaging channels in latent space does not correspond to an interpretable image manipulation. Interestingly, using our method to construct hybrid images with a latent diffusion model, by blurring latent codes, works to an extent but is easily susceptible to artifacts (see \cref{sec:apdx_latent}), so we opt to use a pixel diffusion model which is more consistent and principled.

\subsection{Hybrid Images}
DeepFloyd IF~\cite{deepfloyd2023} generates images in two stages. First at a resolution of $64\times 64$ and then at $256\times 256$. Because of this, we adopt the convention that our $\sigma$ values are specified for the $64\times 64$ scale, and are scaled by $4\times$ for the $256\times 256$ images. We use a relatively large kernel size of $33$ at both scales to minimize edge effects. We use $\sigma$ values ranging from $\sigma=1.0$ to $\sigma=3.0$ for all hybrid images except for those in \cref{fig:sweep}, in which we sweep the value of $\sigma$. 

\subsection{Triple Hybrids}\label{sec:addx_triple}
Triple hybrids are quite difficult to synthesize, and as such we manually select the sigma values and prompts to generate high-quality samples. Specifically, we use $\sigma_1$ values from $\sigma_1=0.8$ to $\sigma_1=1.0$ and $\sigma_2$ values from $\sigma_2=1.2$ to $\sigma_2=2.0$ for all triple hybrids in~\cref{fig:teaser} and \cref{fig:sup_triple}.

\subsection{Upscaling}

DeepFloyd IF additionally uses a third stage which upscales from $256\times 256$ to $1024\times 1024$. We also use this stage, but because it is a latent model, we do not apply our method. We upscale using only the prompt corresponding to the highest frequency component or the color component.

\section{Human Studies}\label{sec:apdx_human_impl}
We use Amazon Mechanical Turk for the human study. 77 ``master workers'' were asked the following questions for each hybrid image pair:

\begin{itemize}[label=$\bullet$]
    \item ``Which image shows \texttt{[prompt\_1]} clearer?'' 
    \item ``Which image shows \texttt{[prompt\_2]} clearer?''
    \item ``Which image is of higher quality?''
\end{itemize}

For low frequency prompt questions, we downsample the images accordingly in order to help participants more easily see the content. For the high frequency prompt questions, as well as the quality questions, we display the images at full resolution. Participants were shown 8 hybrid image pairs in a random order.

\begin{figure}[h]
    \centering
    \includegraphics[width=\linewidth]{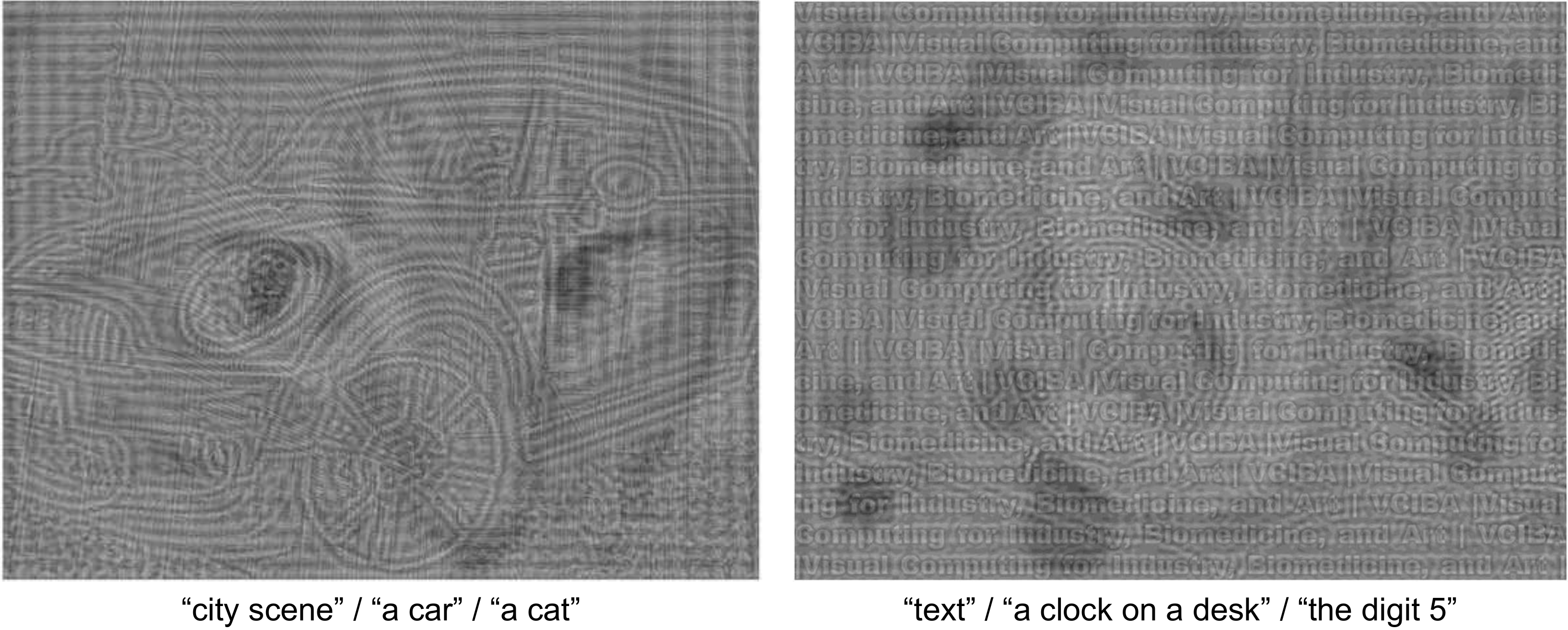}
    \caption{{\bf Prior Work on Triple Hybrid Images.} We show the triple hybrid results from prior work~\cite{sripian2020hybrid}, which adapts the classic method of~\cite{oliva2006hybrid}. A description of what should be seen is provided underneath each image, going from high to low frequencies. As can be seen, these results are of lower quality than our results.}
\label{fig:prior_triple}
\end{figure}
\section{Prior Triple Hybrid Methods}\label{sec:apdx_prior_triple}
Prior work~\cite{sripian2020hybrid} attempts to create triple hybrid images by adapting the method of Oliva~\etal~\cite{oliva2006hybrid}. As can be seen in \cref{fig:prior_triple}, the results are not of high visual quality, and it can be hard to identify the three different subjects in the image, especially when compared to our results. This reflects the difficulty of creating these images.

\section{Metrics Implementation}\label{sec:apdx_metrics}
In \cref{tbl:hybrid_clip}, we report the max CLIP score over multiple image downsampling factors. Specifically, for each hybrid image we downsample and then upsample by a factor $f$, where we choose $f$ to be a linear sweep of 20 values between 1 and 8. These images are then preprocessed to a size of $224 \times 224$, which is the input resolution of the CLIP ViT-B/32 model which we use. We then take the normalized dot product between each resulting image embedding, and the text embedding for the corresponding prompt, and report the max. We report the max to account for the fact that different hybrid images are best seen at different downsampling factors.

\section{Connection to MultiDiffusion}\label{sec:apdx_multidiffusion}

In \cref{sec:multidiffusion} we explore Factorized Diffusion with a spatial decomposition, and show that it allows targeting of prompts to specific spatial regions. We claim that this is a special case of MultiDiffusion~\cite{bar2023multidiffusion}. MultiDiffusion updates a noisy image of arbitrary size by removing the consensus of multiple noise estimates over the image. Factorized Diffusion, with a spatial decomposition, also removes a consensus of multiple noise estimates. However, in our setup this consensus is formed specifically by the disjoint union of multiple noise estimates, and our method operates only at the resolution for which the diffusion model is trained, as opposed to MultiDiffusion.

\begin{figure}[t!]
    \centering
    \includegraphics[width=\linewidth]{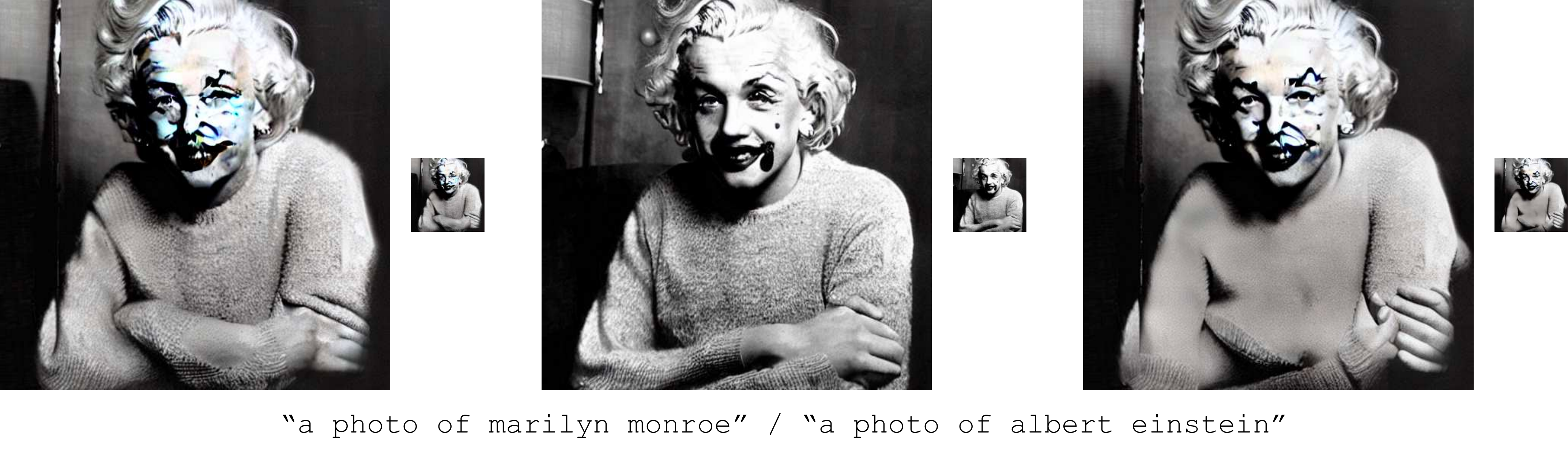}
    \caption{{\bf Latent Hybrid Images.} We provide hybrid image results using our method with Stable Diffusion v1.5, a latent diffusion model. As can be seen the results are passable, but suffer from artifacts, due to applying blurring and bandpass operations in the latent space.}
\label{fig:latent_blur}
\end{figure}
\section{Hybrid Images with Latent Diffusion Models}\label{sec:apdx_latent}
We show hybrid images resulting from using our method with Stable Diffusion v1.5, a latent diffusion model, in \cref{fig:latent_blur}. As can be seen the results are recognizable, but have significant artifacts, due to applying bandpass filters in the latent space. We find that pixel diffusion models produce much higher quality samples.

\section{Synthesizing Hybrid Images with other Methods}\label{sec:illusion_baselines}

\begin{figure}[t!]
    \centering
    \includegraphics[width=\linewidth]{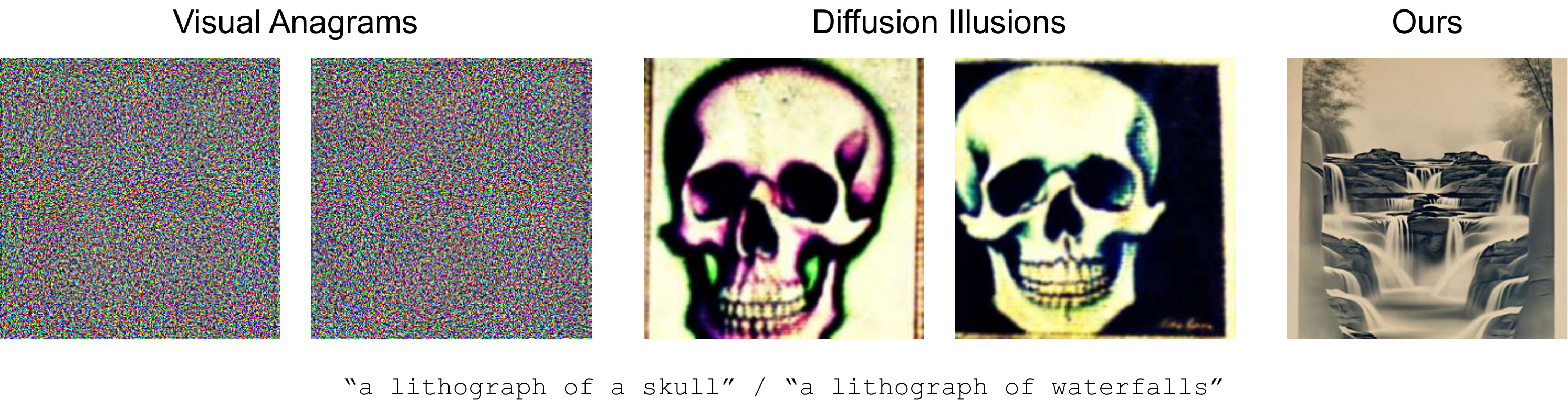}
    \caption{{\bf Other Illusion Methods.} We attempt to create hybrid images using Visual Anagrams~\cite{geng2024visual} and Diffusion Illusions~\cite{burgert2023illusions}, two recent methods designed to generate optical illusions. As can be seen, both methods fail. Please see \cref{sec:illusion_baselines} for analysis.}
\label{fig:illusion_baselines}
\end{figure}

We also attempt to generate hybrid images using two recent methods: Visual Anagrams~\cite{geng2024visual} and Diffusion Illusions~\cite{burgert2023illusions}. Results can be seen in \cref{fig:illusion_baselines}. Both methods fail, which we describe and analyze below.

Diffusion Illusions works by minimizing an SDS~\cite{poole2022dreamfusion} loss over multiple views of an image, paired with different prompts. We use the same high and low pass views as above. As can be seen in \cref{fig:illusion_baselines} the method produces a decent version of the low pass prompt, but fails to incorporate any of the high pass prompt. We believe this is because taking the high pass of an image moves it significantly out-of-distribution, rendering the SDS gradients unhelpful. Low passing an image alters its appearance, but keeps it relatively in-distribution, so as a result the method can still produce the low pass prompt.

Visual Anagrams works by denoising multiple transformations of an image, paired with different prompts. We use a high pass and low pass transformation, but this fails because these operations change the statistics of the noise in the noisy image. As a result, the diffusion model is being fed out-of-distribution images, and the reverse process fails to converge, as shown in \cref{fig:illusion_baselines}.

\begin{table}[t!]
    \begin{center}
    \caption{{\bf Comparison to Visual Anagrams~\cite{geng2024visual}.} We use \cite{geng2024visual} to synthesize hybrids and color hybrids, and report the same metrics as \cite{geng2024visual}. We use prompt pairs built from the CIFAR-10 classes, with 10 prompts per pair for a total of 900 samples. Our method performs consistently better, as~\cite{geng2024visual} is not designed to produce these kinds of illusions.}
    
    \label{tbl:baselines}
    \setlength\tabcolsep{3pt}
    \resizebox{0.8\linewidth}{!}{
    \begin{tabular}{llcccccc}
    \toprule
    Task & Method
    & $\mathcal{A}\uparrow$ & $\mathcal{A}_{0.9}\uparrow$ & $\mathcal{A}_{0.95}\uparrow$ & $\mathcal{C}\uparrow$ & $\mathcal{C}_{0.9}\uparrow$ & $\mathcal{C}_{0.95}\uparrow$ \\
    \midrule
    
    \multirow{ 2}{*}{Hybrid Images} & Visual Anagrams~\cite{geng2024visual} & 0.226 & 0.237 & 0.240 & 0.500 & 0.520 & 0.525 \\
    & Ours & \textbf{0.237} & \textbf{0.263} & \textbf{0.271} & \textbf{0.536} & \textbf{0.630} & \textbf{0.651} \\
    \multirow{ 2}{*}{Color Hybrids} & Visual Anagrams~\cite{geng2024visual} & 0.223 & 0.232 & 0.234 & 0.500 & 0.537 & 0.547 \\
    & Ours & \textbf{0.231} & \textbf{0.260} & \textbf{0.269} & \textbf{0.512} & \textbf{0.562} & \textbf{0.586} \\
    
    \bottomrule
    \end{tabular}
   }
    \end{center}
    \vspace{-8mm}
\end{table}

Finally, we also quantitatively evaluate hybrid and color hybrids generated using Geng~\etal\cite{geng2024visual} against our proposed method, with results shown in \cref{tbl:baselines}. As prompts, we use all pairs of CIFAR-10 classes, and sample 10 images per prompt pair for a total of 900 samples. We use the same metrics as~\cite{geng2024visual}, and we find that our method does better consistently, as~\cite{geng2024visual} was not designed to generate these illusions.

\section{Further Analysis of Factorized Diffusion}\label{sec:apdx_analysis}

As discussed in \cref{sec:analysis}, our analysis assumes that the update step is a linear combination of the noisy image, $\x_t$, and the noise estimate, $\epsilon_\theta$. However, many commonly used update steps also involve adding random noise $\mathbf{z}\sim\mathcal{N}(0,\mathbf{I})$, such as DDPM~\cite{ho2020denoising}. To deal with this, we can view the update step as a composition of two steps:
\begin{align}
    \x_{t-1} &= \texttt{update}(\x_t, \epsilon_\theta) \\
    &= \texttt{update'}(\x_t, \epsilon_\theta) + \sigma_z\mathbf{z}.
\end{align}
The first step is a linear combination of $\x_t$ and $\epsilon_\theta$, and the second adds in the noise $\mathbf{z}$. Our analysis then applies to just the $\texttt{update'}$ function.

\section{Choosing Prompts}\label{sec:apdx_prompts}
We find that carefully choosing prompts can generate higher quality illusions. For example, the success rate and quality of samples are much higher when at least one prompt is of a ``flexible'' subject, such as \texttt{"houseplants"} or \texttt{"a canyon"}. In addition, we found biases specific to decompositions. Prompts with the style \texttt{"photo of..."} performed better for hybrid and motion hybrid images. We suspect this is because photos tend to have ample amounts of both high and low frequency content, as opposed to styles such as \texttt{"oil paintings"} or \texttt{"watercolors"}, which tend to lack higher frequency content. For color hybrids, we found that using the style of \texttt{"watercolor"} produced better results, perhaps because of the style's emphasis on color.

\begin{figure}[h]
    \centering
    \vspace{-1em}
    \includegraphics[width=0.9\linewidth]{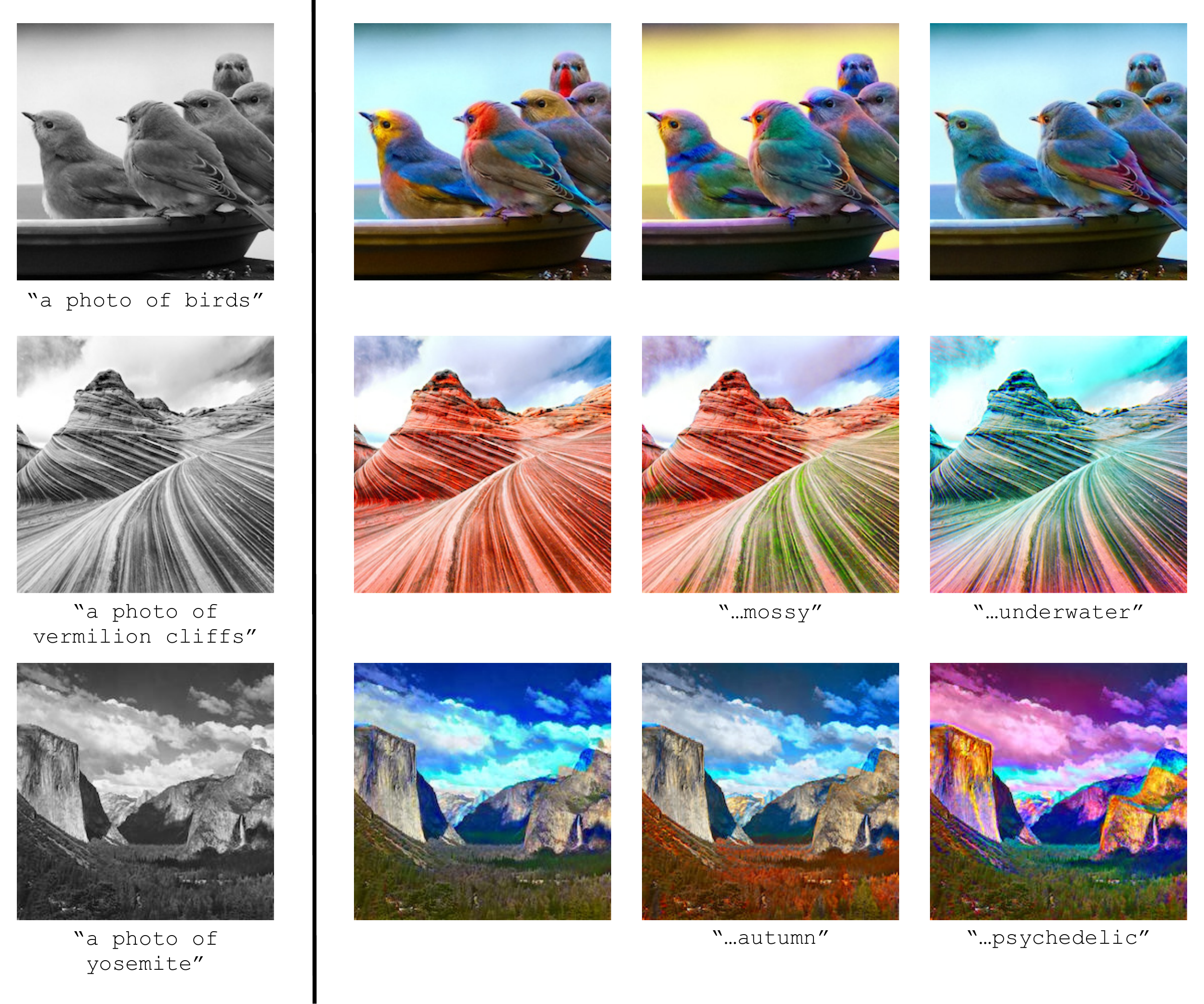}
    \caption{{\bf Colorization.} Our method can also be used to solve inverse problems, such as colorization. We show grayscale images that we wish to colorize on the left. The color component is then generated conditioned on the text prompts displayed. Note that this is effectively prior work~\cite{song2021scorebased,kawar2022denoising,chung2022improving}.}
    \vspace{-2em}
\label{fig:inverse_colorization}
\end{figure}
\section{Colorization}\label{sec:apdx_colorization}

We also show colorization results in \cref{fig:inverse_colorization}, using our method as an inverse problem solver, as discussed in \cref{sec:method_inverse}. Specifically, we use the color space decomposition introduced in \cref{sec:decompositions_considered}. During diffusion model sampling we hold the grayscale component fixed to the grayscale component of a real image that we want to colorize, and generate the color component. Note that this is effectively prior work~\cite{song2021scorebased,kawar2022denoising,chung2022improving}.

\section{Additional Results}\label{sec:apdx_more_results}
\begin{figure}[h]
    \centering
    \includegraphics[width=\linewidth]{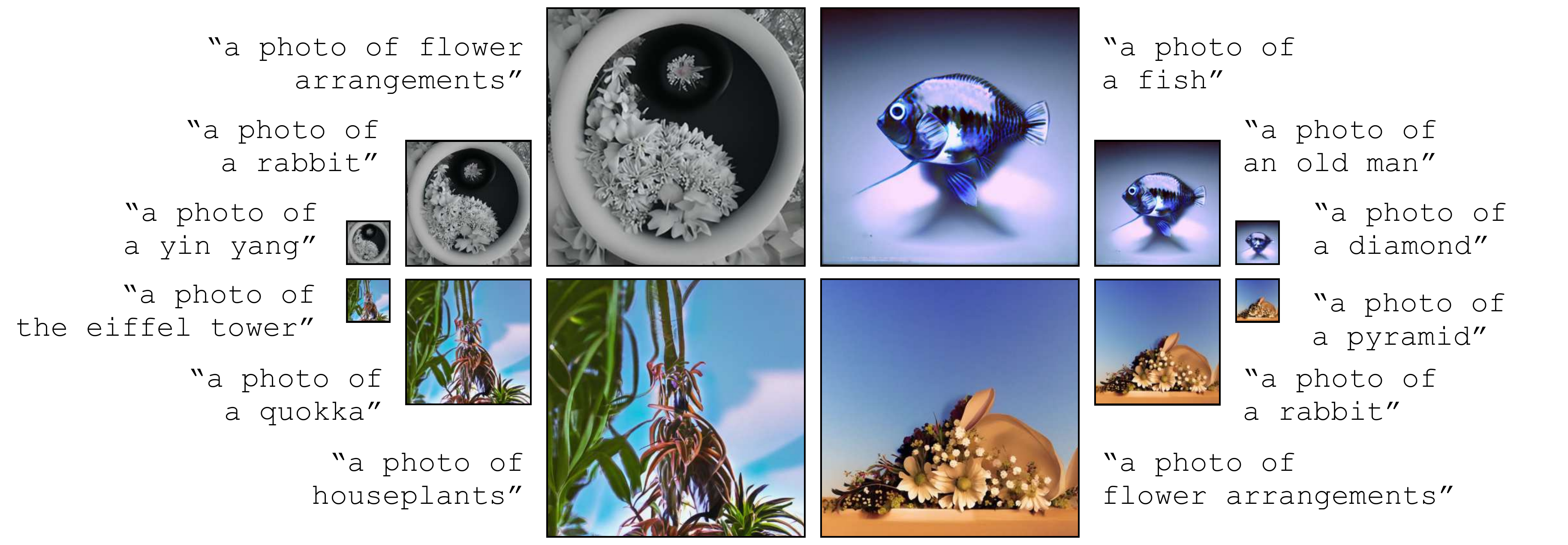}
    \caption{{\bf Triple Hybrids.} We provide more {\it triple hybrid} results. \textbf{\textit{Best viewed digitally, using zoom.}}}
\label{fig:sup_triple}
\end{figure}

\begin{figure}[h]
    \centering
    \includegraphics[width=\linewidth]{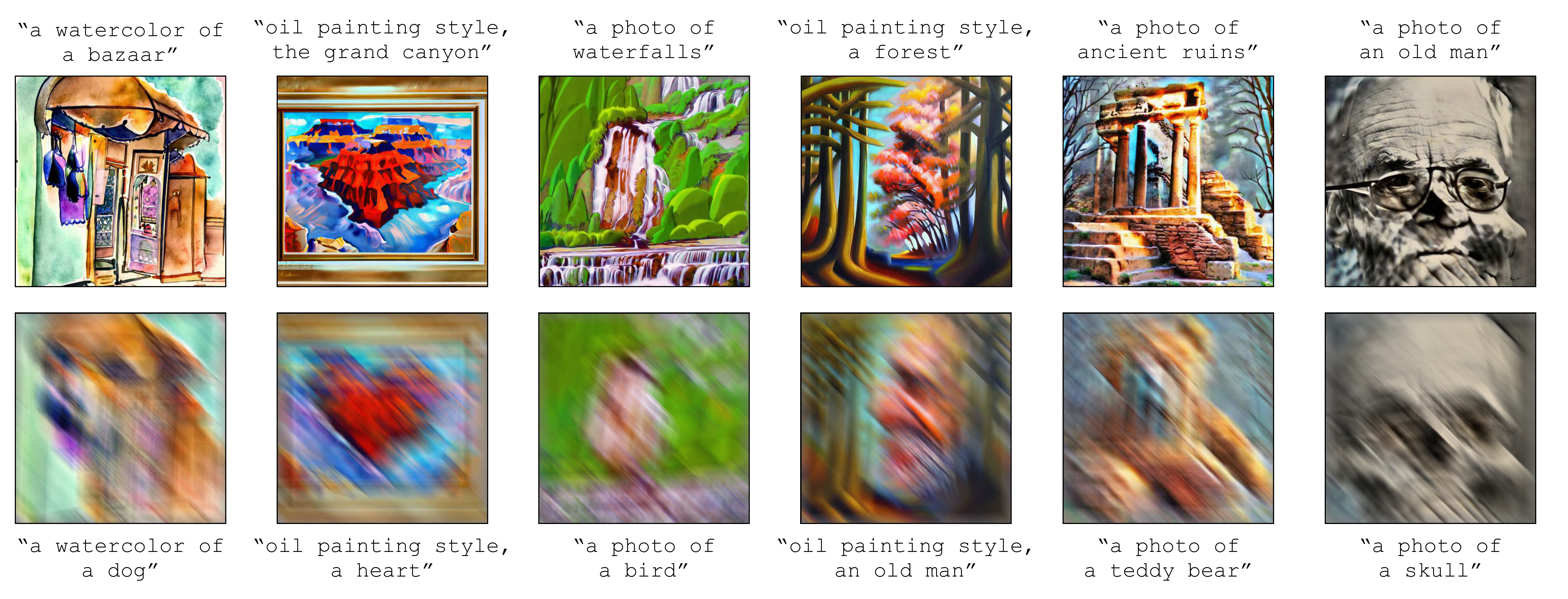}
    \caption{{\bf Motion Hybrids.} We show more {\it motion hybrid} results. These are images that change appearance when motion blurred. Here, the motion is from upper left to bottom right.}
\label{fig:sup_motion}
\end{figure}

\begin{figure}[h]
    \centering
    \includegraphics[width=\linewidth]{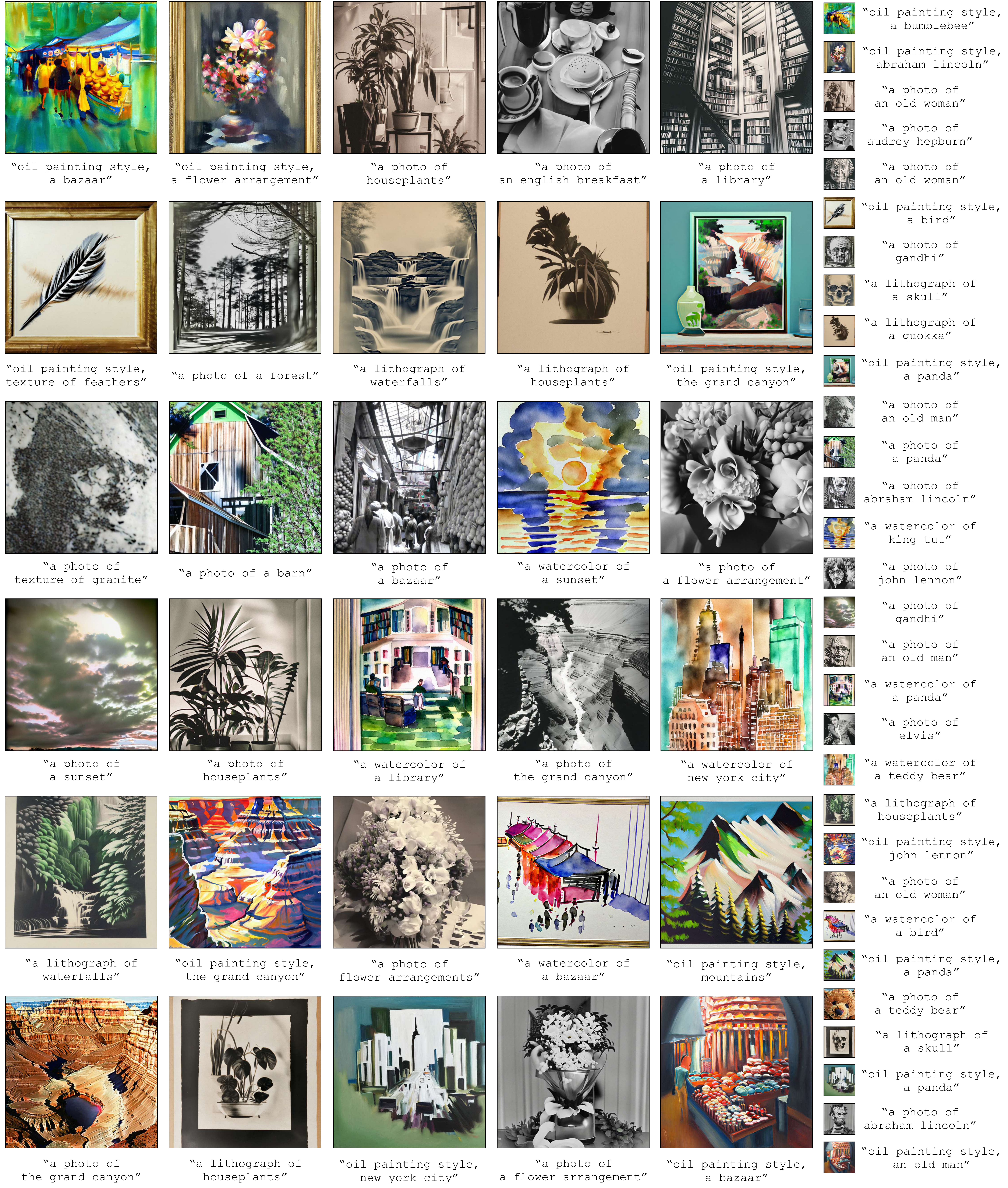}
    \caption{{\bf Hybrid Images.} We show more {\it hybrid image} results. For easier viewing, we provide insets of each hybrid image at lower resolution, along with the corresponding prompt. \textbf{\textit{Best viewed digitally, with zoom.}}}
\label{fig:sup_hybrid}
\end{figure}

\begin{figure}[h]
    \centering
    \includegraphics[width=\linewidth]{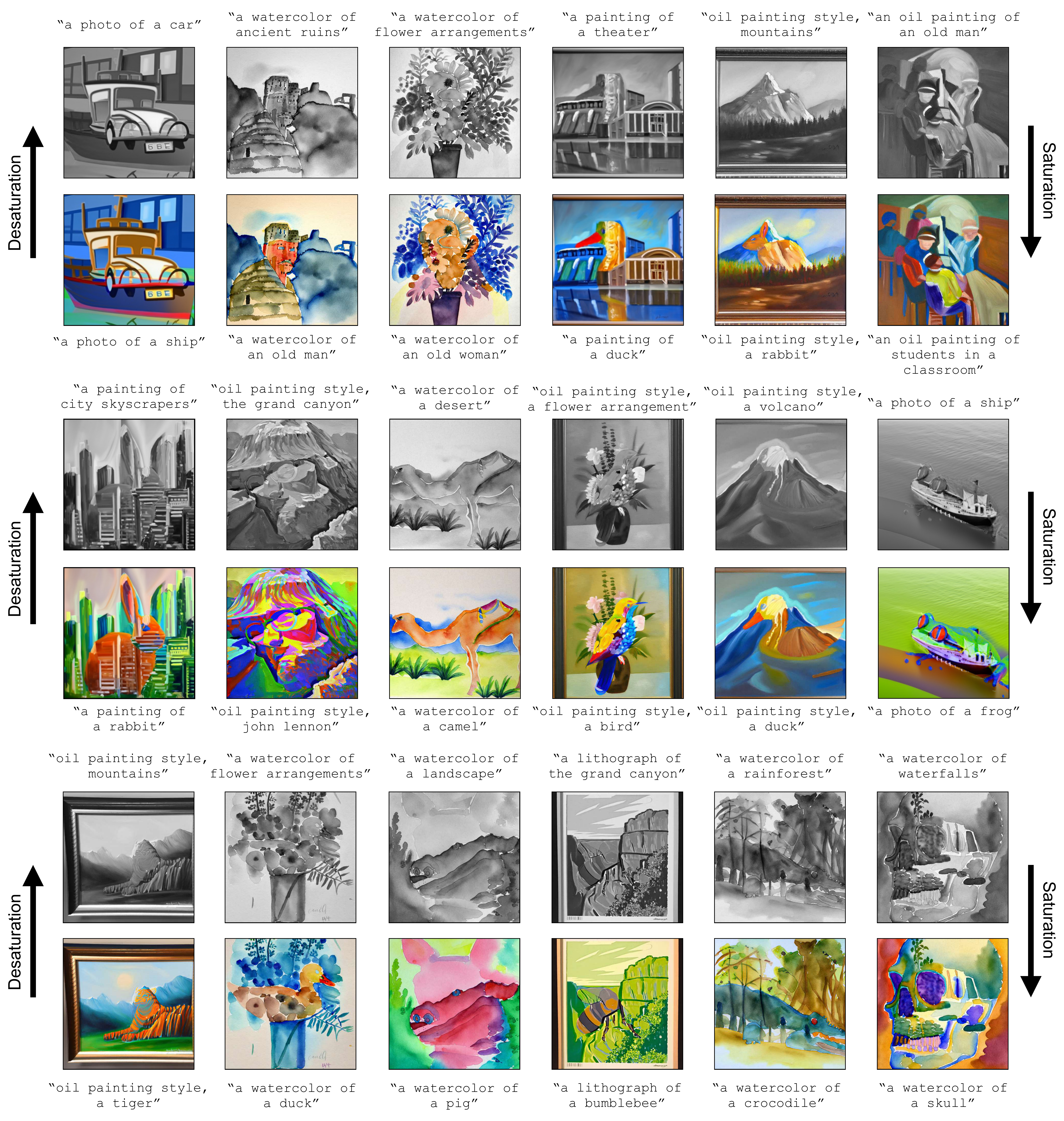}
    \caption{{\bf Color Hybrids.} We show more {\it color hybrid} results, with grayscale images placed above their colorized version.}
\label{fig:sup_color}
\end{figure}

\begin{figure}[h]
    \centering
    \includegraphics[width=\linewidth]{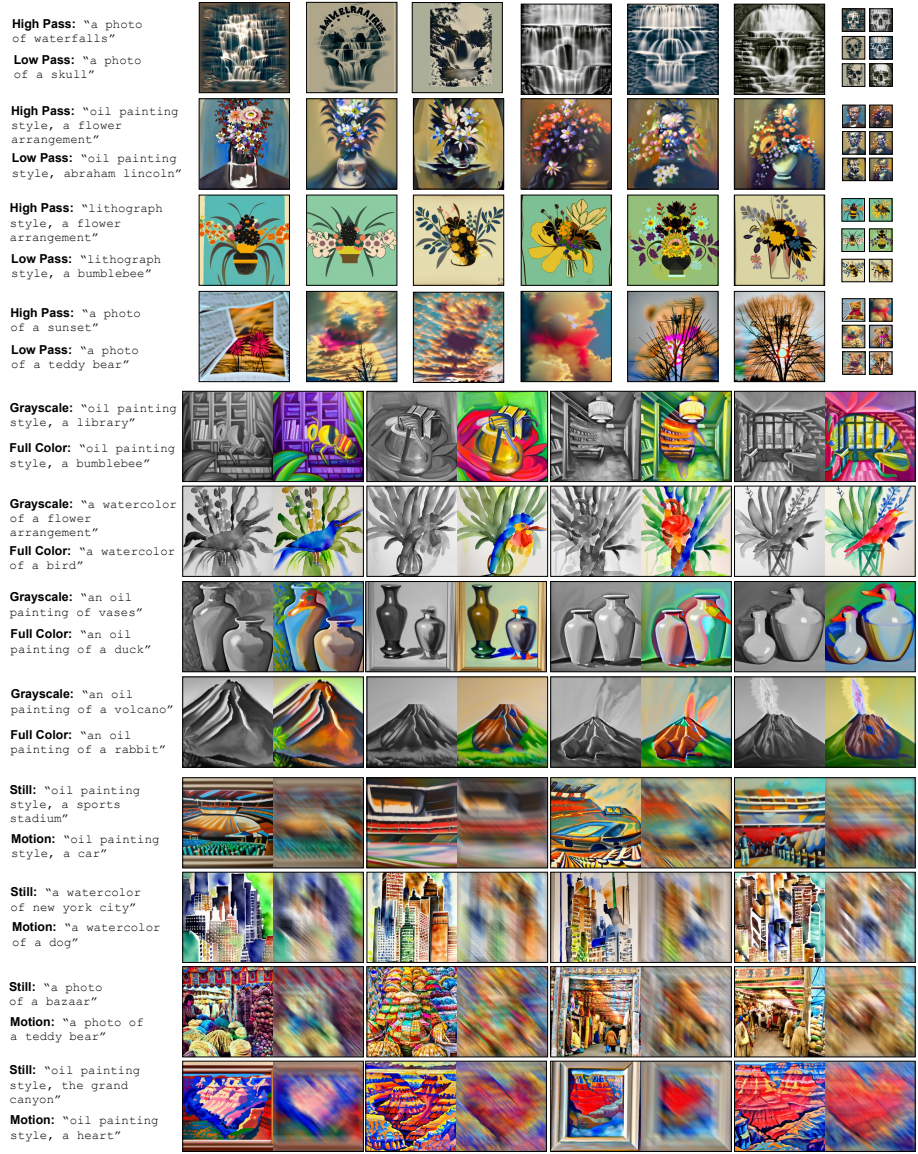}
    \caption{{\bf Random Samples.} We provide random samples of hybrid images, color hybrids, and motion hybrids for selected prompts.
    }
\label{fig:sup_random}
\end{figure}

In this section, we provide additional qualitative results. Additional results for hybrid images and triple hybrids are shown in \cref{fig:sup_hybrid} and \cref{fig:sup_triple} respectively. In \cref{fig:sup_motion} and \cref{fig:sup_color}, we provide more examples of motion and color hybrids, respectively. Finally, we provide more random samples for hybrid images, color hybrids, and motion hybrids in \cref{fig:sup_random}.


\clearpage
\section{CVPR 2024 T-Shirt Design}\label{sec:apdx_cvpr_tshirt}

We created an inverse hybrid image for the official CVPR 2024 T-shirt as part of the AI Art track. Our goal was for attendees to see only a watercolor of the Seattle skyline when they received the shirt. Then, as they see other people wearing the shirts in the conference center from a distance, the text ``CVPR'' would be revealed.  

\begin{figure}[h]
    \centering
    \includegraphics[width=\linewidth]{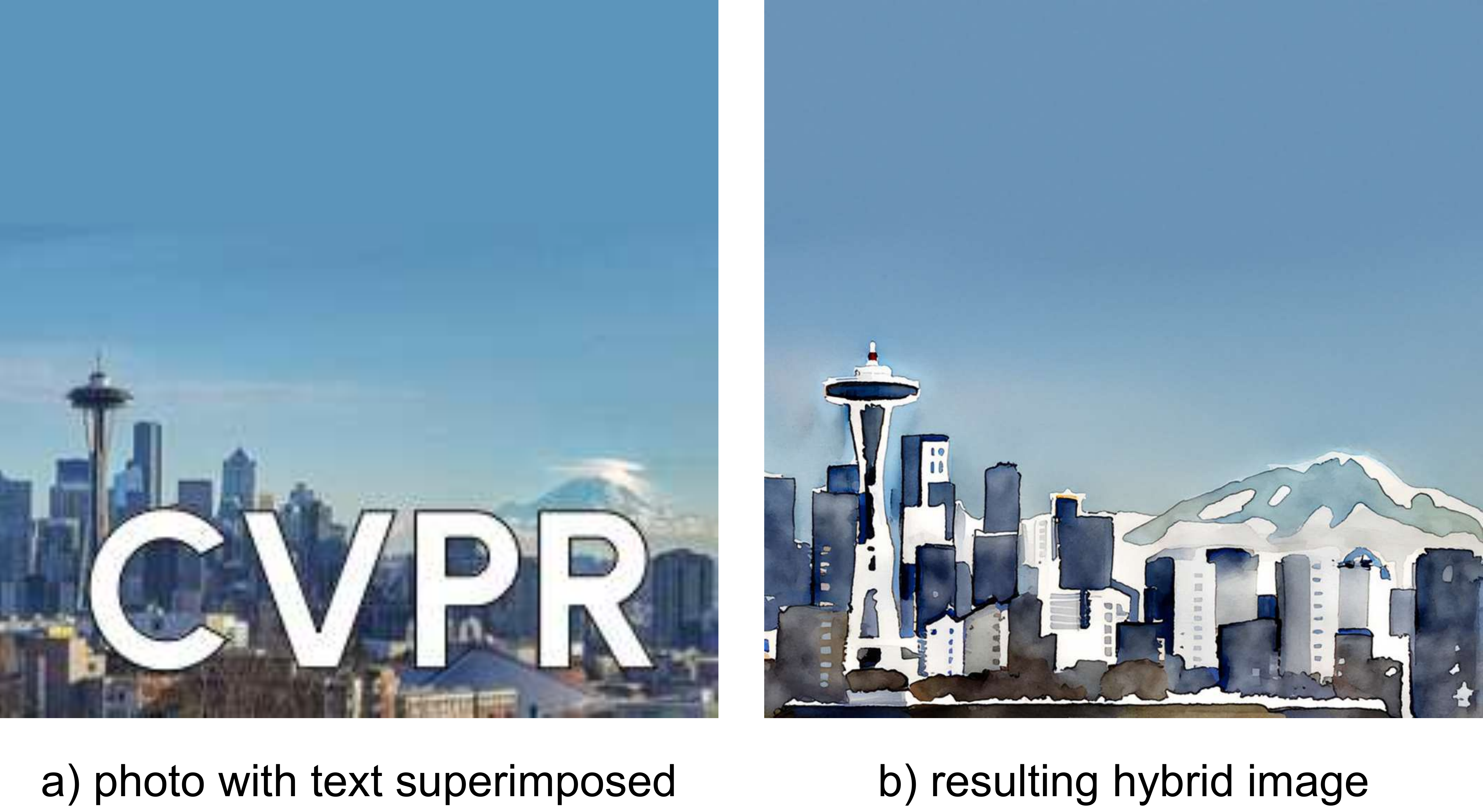}
    \caption{{\bf CVPR Hybrid Image T-shirt Design.} \textbf{a)} The edited photo, from which we extract low spatial frequencies. \textbf{b)} The resulting hybrid image, after generating high spatial frequencies conditioned on the extracted low frequencies. For more details, please visit our \href{https://dangeng.github.io/factorized_diffusion/}{website}. Photo source: Pavol Svantner~\cite{svantner2024seattle}.}  
\label{fig:cvpr_tshirt}
\end{figure}

\begin{figure}[h]
    \centering
    \includegraphics[width=\linewidth]{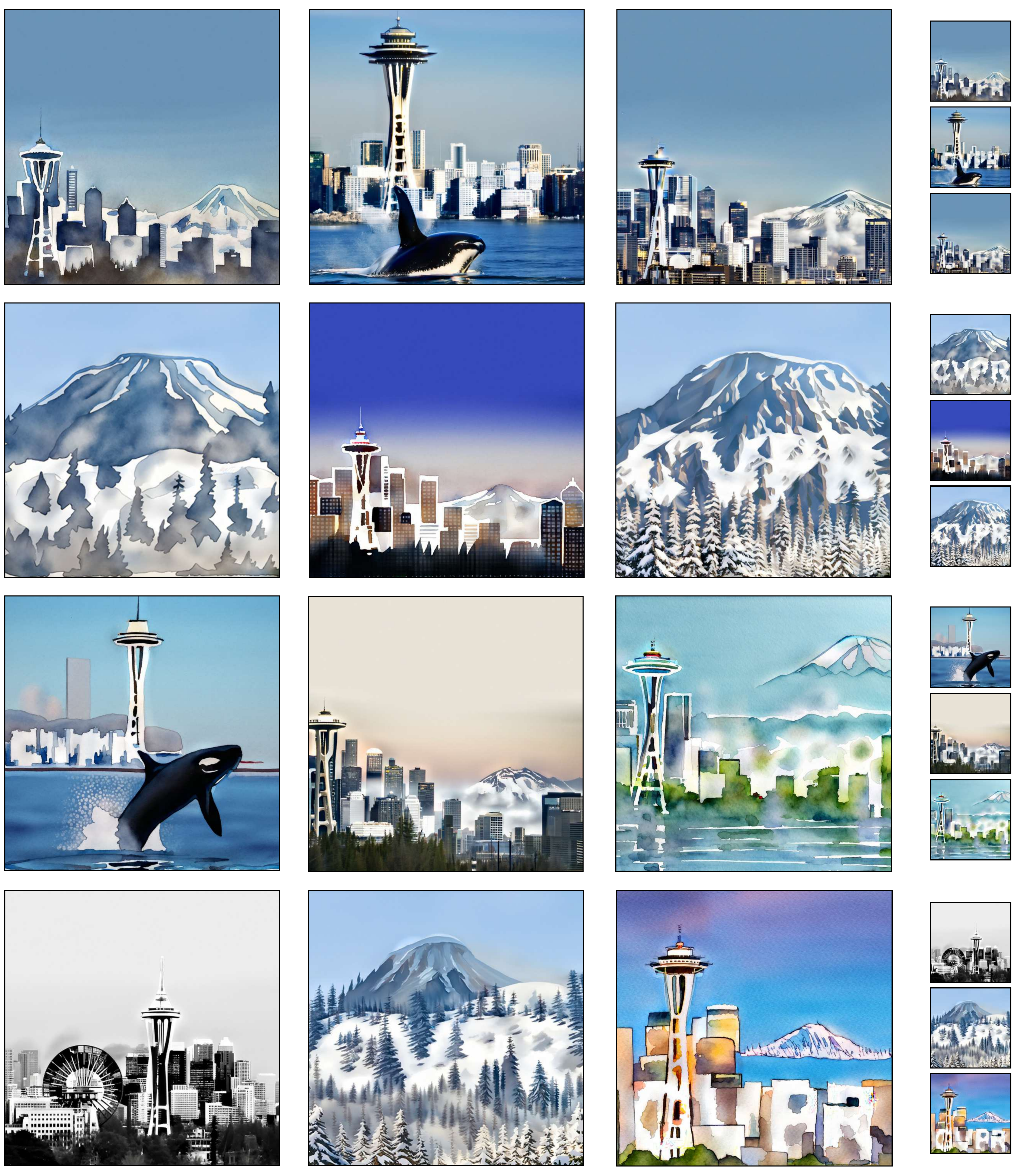}
    \caption{{\bf CVPR Hybrid Image T-shirt Design Candidates.} We show more CVPR T-shirt designs. For easier viewing, we provide insets of each hybrid image at lower resolution. \textbf{\textit{Best viewed digitally, with zoom.}}}  
\label{fig:cvpr_grid}
\end{figure}

We took an existing photo of the Seattle skyline, and pasted the text ``CVPR'' over the image. We then used our technique to condition an image on the low frequencies of the edited photo, and fill in the high frequencies given the text prompt \texttt{``a watercolor of the seattle skyline with mount rainier in the background''}. The resulting image was then touched up by running Adobe Photoshop's generative fill in a few locations with artifacts to improve quality. We show the low frequency image and the hybrid image, before editing, in \cref{fig:cvpr_tshirt}. We also show additional candidate T-shirt designs in \cref{fig:cvpr_grid}, which all reveal the text ``CVPR'' when viewed from a distance.


\end{document}